\documentclass{article}

\usepackage{FRtemplates/frExamplee}
\usepackage{FRtemplates/apalike}
\let\bibhang\relax
\usepackage{blindtext}
\usepackage{graphicx}
\usepackage{setspace}
\usepackage{color,soul}
\usepackage{subfigure}
\usepackage{colortbl}
\usepackage{enumitem}
\usepackage{tikz}
\usepackage{pgfplots}
\usepackage{tkz-euclide}
\usepackage{amsmath}
\usepackage{bm}
\usepackage{amssymb}
\usepackage{moreverb,url}
\usepackage{textgreek}
\usepackage[acronym,toc]{glossaries}
\usepackage{wrapfig}
\usepackage{lipsum} 
\usepackage{multirow}
\usepackage{placeins}
\usepackage[colorlinks,bookmarksopen,bookmarksnumbered,citecolor=blue,urlcolor=blue]{hyperref}
\usepackage{lipsum}
\usepackage{mathptmx}
\usepackage{latexsym}
\usepackage{graphics}
\usepackage{epsfig}
\usepackage[english]{babel}
\usepackage{natbib}
\usepackage{times}
\usepackage[linesnumbered,lined,ruled,commentsnumbered]{algorithm2e}
\usepackage{algpseudocode}
\usepackage{amssymb}
\usepackage{epstopdf}
\usepackage{lineno}
\usepackage{multicol}
\usepackage{bbm}
\usepackage{microtype}
\usepackage{booktabs}
\usepackage{caption}
\usepackage{txfonts}
\usepackage{floatrow}
\usepackage{authblk}
\usepackage{color}
\usepackage{xcolor}
\usepackage[export]{adjustbox}
\usepackage{pifont}

\newcommand{\cmark}{\ding{51}}%
\newcommand{\xmark}{\ding{55}}%

\newcommand\blue[1]{{}{#1}} 
\newcommand\green[1]{{}{#1}} 
\newcommand\red[1]{{}{#1}} 

\AppendGraphicsExtensions{.tiff}
\AppendGraphicsExtensions{.tif}

\definecolor{light-blue}{RGB}{174,214,241} 			
\definecolor{middle-light-blue}{RGB}{52,152,219} 	
\definecolor{dark-blue}{RGB}{26,76,110} 			
\definecolor{very-light-gray}{RGB}{230,231,232}		
\definecolor{light-gray}{RGB}{171,178,185}			
\definecolor{middle-light-gray}{RGB}{86,101,115}	
\definecolor{dark-gray}{RGB}{44,62,80}				
\definecolor{light-red}{RGB}{231,76,60}				%
\definecolor{dark-red}{RGB}{173,57,45}				%

\usepackage{amsmath,amsfonts,bm}

\def\eqref#1{equation~\ref{#1}}

\def\1{\bm{1}}

\def\vb{{\bm{b}}}
\def\vc{{\bm{c}}}

\def\vh{{\bm{h}}}

\def\vn{{\bm{n}}}

\def\vp{{\bm{p}}} 
\def\vq{{\bm{q}}}

\def\vt{{\bm{t}}} 

\def\vx{{\bm{x}}}
\def\vy{{\bm{y}}}

\def\mC{{\bm{C}}}

\def\mH{{\bm{H}}} 
\def\mI{{\bm{I}}}

\def\mP{{\bm{P}}}
\def\mQ{{\bm{Q}}}
\def\mR{{\bm{R}}} 

\def\mT{{\bm{T}}} 
\def\mU{{\bm{U}}}
\def\mV{{\bm{V}}}

\DeclareMathAlphabet{\mathsfit}{\encodingdefault}{\sfdefault}{m}{sl}
\SetMathAlphabet{\mathsfit}{bold}{\encodingdefault}{\sfdefault}{bx}{n}

\newcommand{\Ls}{\mathcal{L}}

\newcommand{\data}{\mathcal{D}}
\newcommand{\batch}{\mathcal{B}}
\newcommand{\inputs}{\mathcal{X}}
\newcommand{\outputs}{\mathcal{Y}}

\newcommand{\thetamap}{{\boldsymbol{\hat{\theta}}}}

\newglossaryentry{VO}{name=VO,description={Visual Odometry},first={Visual Odometry (VO)}}
\newglossaryentry{IMU}{name=IMU,description={inertial measurement unit},first={Inertial Measurement Unit (IMU)}}
\newglossaryentry{FOV}{name=FOV,description={field of view},first={field of view (FOV)}}
\newglossaryentry{RAFCON}{name=RAFCON,description={RMC Advanced Flow Control},first={RMC Advanced Flow Control (RAFCON)}}
\newglossaryentry{IAC}{name=IAC,description={International Astronautical Congress},first={International Astronautical Congress (IAC)}}
\newglossaryentry{VR}{name=VR,description={Virtual Reality},first={Virtual Reality (VR)}}
\newglossaryentry{AR}{name=AR,description={Augmented Reality},first={Augmented Reality (AR)}}
\newglossaryentry{GNSS}{name=GNSS,description={Global Navigation Satellite System},first={Global Navigation Satellite System (GNSS)}}
\newglossaryentry{VINS}{name=VINS,description={Visual Inertial System},first={Visual Inertial System (VINS)}}
\newglossaryentry{COG}{name=CoG,description={Center of Gravity},first={Center of Gravity (CoG)}}
\newglossaryentry{SAR}{name=SAR,description={Search and Rescue},first={Search and Rescue (SAR)}}
\newglossaryentry{FPGA}{name=FPGA,description={Field Programmable Gate Array},first={Field Programmable Gate Array (FPGA)}}
\newglossaryentry{PCB}{name=PCB,description={Printed Circuit Board},first={Printed Circuit Board (PCB)}}
\newglossaryentry{SGM}{name=SGM,description={Semi-Global matching},first={Semi-Global matching (SGM)}}
\newglossaryentry{SPI}{name=SPI,description={Serial Peripheral Interface},first={Serial Peripheral Interface (SPI)}}
\newglossaryentry{DRL}{name=DRL,description={Drone Racing League},first={Drone Racing League (DRL)}}
\newglossaryentry{CAN}{name=CAN-bus,description={Controller Area Network},first={Controller Area Network (CAN-bus)}}
\newglossaryentry{LRU}{name=LRU,description={Lightweight Rover Unit},first={Lightweight Rover Unit (LRU)}}
\newglossaryentry{ARDEA}{name=ARDEA,description={Autonomous Robot Design for Extraterrestrial Applications},first={Autonomous Robot Design for Extraterrestrial Applications (ARDEA)}}
\newglossaryentry{SLAM}{name=SLAM,description={Simultaneous Localization and Mapping},first={Simultaneous Localization and Mapping (SLAM)}}
\newglossaryentry{ROBEX}{name=ROBEX,description={Robotic Exploration of Extreme Environments},first={Robotic Exploration of Extreme Environments (ROBEX)}}
\newglossaryentry{MOSFET}{name=MOSFET,description={MOSFET},first={Metal-Oxide-Semiconductor Field-Effect Transistor, (MOSFET)}}
\newglossaryentry{BBB}{name=BBB,description={BeagleBone Black},first={BeagleBone Black (BBB)}}
\newglossaryentry{SVD}{name=SVD,description={Singular Value Decomposition},first={Singular Value Decomposition (SVD)}}
\newglossaryentry{PD}{name=PD,description={Proportional-Derivative},first={Proportional-Derivative (PD)}}
\newglossaryentry{HLWP}{name=High Level Waypoint Planner ,description={High Level Waypoint Planner },first={High Level Waypoint Planner (HLWP)}}
\newglossaryentry{RT}{name=RT,description={real-time},first={real-time (RT)}}
\newglossaryentry{nonRT}{name=nonRT,description={non-RT},first={non real-time (non-RT)}}
\newglossaryentry{OS}{name=OS,description={operating system},first={operating system (OS)}}
\newglossaryentry{AUTOPILOT}{name=AUTOPILOT,description={AUTOPILOT},first={AUTOPILOT (Automated driving Progressed by Internet Of Things)}}
\newglossaryentry{LIDAR}{name=LIDAR,description={LIDAR},first={Light Detection and Ranging (LIDAR)}}
\newglossaryentry{RADAR}{name=RADAR,description={RADAR},first={Radio Detection and Ranging (RADAR)}}
\newglossaryentry{ROS}{name=ROS,description={Robot Operating System},first={Robot Operating System (ROS)}}
\newglossaryentry{PTPd}{name=PTPd,description={Precision Time Protocol daemon},first={Precision Time Protocol daemon (PTPd)}}
\newglossaryentry{SDA}{name=SDA,description={Strapdown Algorithm},first={Strapdown Algorithm (SDA)}}
\newglossaryentry{PWM}{name=PWM,description={pulse width modulation},first={pulse width modulation (PWM)}}
\newglossaryentry{MEMS}{name=MEMS,description={micro electro mechanical system},first={micro electro mechanical system (MEMS)}}
\newglossaryentry{TRL}{name=TRL,description={Technology Readiness Level},first={Technology Readiness Level (TRL)}}
\newglossaryentry{NED}{name=NED,description={North East Down},first={North East Down (NED)}}
\newglossaryentry{NDI}{name=NDI,description={Nonlinear Dynamic Inversion},first={Nonlinear Dynamic Inversion (NDI)}}
\newglossaryentry{INDI}{name=INDI,description={Incremental Nonlinear Dynamic Inversion},first={Incremental Nonlinear Dynamic Inversion (INDI)}}
\newglossaryentry{AGAST}{name=AGAST,description={Adaptive and Generic Accelerated Segment Test},first={Adaptive and Generic Accelerated Segment Test (AGAST)}}
\newglossaryentry{EKF}{name=EKF,description={Extended Kalman Filter},first={Extended Kalman Filter (EKF)}}
\newacronym[plural={RRTs}, longplural={Rapidly-Exploring Random Trees}]{RRT}{RRT}{Rapidly-Exploring Random Tree}
\newacronym[plural={ESCs}, longplural={Electronic Speed Controllers}]{ESC}{ESC}{Electronic Speed Controller}
\newacronym[plural={APIs}, longplural={Application Programming Interfaces}]{API}{API}{Application Programming Interface}
\newacronym[plural={DOFs}, longplural={Degrees of Freedom}]{DOF}{DOF}{Degrees of Freedom}
\newacronym[plural={VIOs}, longplural={Visual Inertial Odometries}]{VIO}{VIO}{Visual Inertial Odometry}
\newacronym{JPL}{JPL}{Jet Propulsion Laboratory}
\newacronym[plural={POIs}, longplural={Points of Interest}]{POI}{POI}{Point of Interest}
\newacronym[plural={MAVs}, longplural={Micro Aerial Vehicles}]{MAV}{MAV}{Micro Aerial Vehicle}

\newcommand{\sectionref}[1]{Section~\ref{#1}}

\SetCommentSty{mycommfont}

\title{Virtual Reality via Object Pose Estimation and \\ Active Learning: Realizing Telepresence Robots \\ with Aerial Manipulation Capabilities}

\author[1,2,*]{\textbf{Jongseok~Lee}}
\author[1]{\textbf{Ribin~Balachandran}}
\author[1]{\textbf{Konstantin~Kondak}}
\author[1,3]{\textbf{Andre~Coelho}}
\author[1]{\textbf{Marco~~De~Stefano}}
\author[1]{\\\textbf{Matthias~Humt}}
\author[1]{\textbf{Jianxiang~Feng}}
\author[2]{\textbf{Tamim~Asfour}}
\author[1,4]{\textbf{Rudolph~Triebel}}
\affil[1]{Institute of Robotics and Mechatronics, German Aerospace Center (DLR)}
\affil[2]{Institute for Anthropomatics and Robotics, Karlsruhe Institute of Technology (KIT)}
\affil[3]{Robotics and Mechatronics Laboratory, University of Twente (UT)}
\affil[4]{Chair of Computer Vision and Artificial Intelligence, Technical University of Munich (TUM)}
\affil[*]{Correspondence to jongseok.lee@dlr.de}

\begin{document}

\maketitle

\begin{abstract}
This article presents a novel telepresence system for advancing aerial manipulation in dynamic and unstructured environments. The proposed system not only features a haptic device, but also a virtual reality (VR) interface that provides real-time 3D displays of the robot's workspace as well as a haptic guidance to its remotely located operator. To realize this, multiple sensors namely a LiDAR, cameras and IMUs are utilized. For processing of the acquired sensory data, pose estimation pipelines are devised for industrial objects of both known and unknown geometries. We further propose an active learning pipeline in order to increase the sample efficiency of a pipeline component that relies on Deep Neural Networks (DNNs) based object detection. All these algorithms jointly address various challenges encountered during the execution of perception tasks in industrial scenarios. In the experiments, exhaustive ablation studies are provided to validate the proposed pipelines. Methodologically, these results commonly suggest how an awareness of the algorithms' own failures and uncertainty (`introspection') can be used tackle the encountered problems. Moreover, outdoor experiments are conducted to evaluate the effectiveness of the overall system in enhancing aerial manipulation capabilities. In particular, with flight campaigns over days and nights, from spring to winter, and with different users and locations, we demonstrate over 70 robust executions of pick-and-place, force application and peg-in-hole tasks with the DLR cable-Suspended Aerial Manipulator (SAM). As a result, we show the viability of the proposed system in future industrial applications\footnote{A video material accompanying this paper can be found at \url{https://www.youtube.com/watch?v=JRnPIARW8xY}}.

\end{abstract}

\section{Introduction}
\label{sec:introduction}

The global market for robotic inspection and maintenance is growing fast with an expected annual turnover of up to 4.37 billion dollars by 2025\footnote{BIS Research, Global Inspection and Maintenance Robot Market: Focus on Type, Component, and End User - Analysis and Forecast, 2020-2025; March 2020}. Recently, international corporations and organizations, such as General Electric, Sprint Robotics, Baker Hughes and Boston Dynamics, have started initiatives to generate and evaluate robotic technologies for inspection and maintenance applications. One of the most prominent directions for these real world industrial applications is  aerial manipulation  \citep{ollero2021past}. \red{An aerial manipulation system is composed of robotic manipulators and a controlled flying platform \citep{fishman2021dynamic, bodie2020active, Kondak2014, Kim2013}.} The platform enables coarse positioning while the manipulator enables dexterous grasping and manipulation for complex tasks. Hence, these aerial platforms  extend the mobility of  robotic manipulators, which can be deployed at high altitudes above ground, increasing  safety for  human workers while reducing  costs. Examples of aerial manipulation applications range from load transportation \citep{Bernard2009}, contact based inspection and maintenance in chemical plants \citep{Angel2019}, bridges \citep{Cuevas2019}, power-line maintenance  \citep{cacace2021safe}, to sensor installations in  forests for fire prevention \citep{hamaza2019compact}. 

\red{In this article, the real world applications of aerial manipulators are envisioned for several industrial scenarios in dynamic and unstructured  environments.} For these industrial applications of aerial manipulators, our current interests are in the bilateral teleoperation concepts, i.e., a human operator remotely controls the robotic manipulator from a safe area on ground and receives visual and haptic feedback from the robot. This  increases  human operator safety while the robots execute their tasks in dangerous environments \citep{hulin2021model, Hirzinger2003}. Such a concept is motivated  by having a robotic system  with a human-in-the-loop, where the system can leverage human intelligence to reliably accomplish its missions. To realize this, existing works have focused on relevant components of the system, namely force feedback teleoperation under time delays \citep{balachandran2021stabilization, spacejoystick}, shared autonomy \citep{masone2018shared}, human-machine interfaces \citep{kim2021toward, yashin2019aerovr, wu2018aerial}, and robotic perception for aerial manipulators \citep{Karrer2016, pumarola2019relative}.

\begin{figure}
 \centering
 \includegraphics[width=1.0\textwidth]{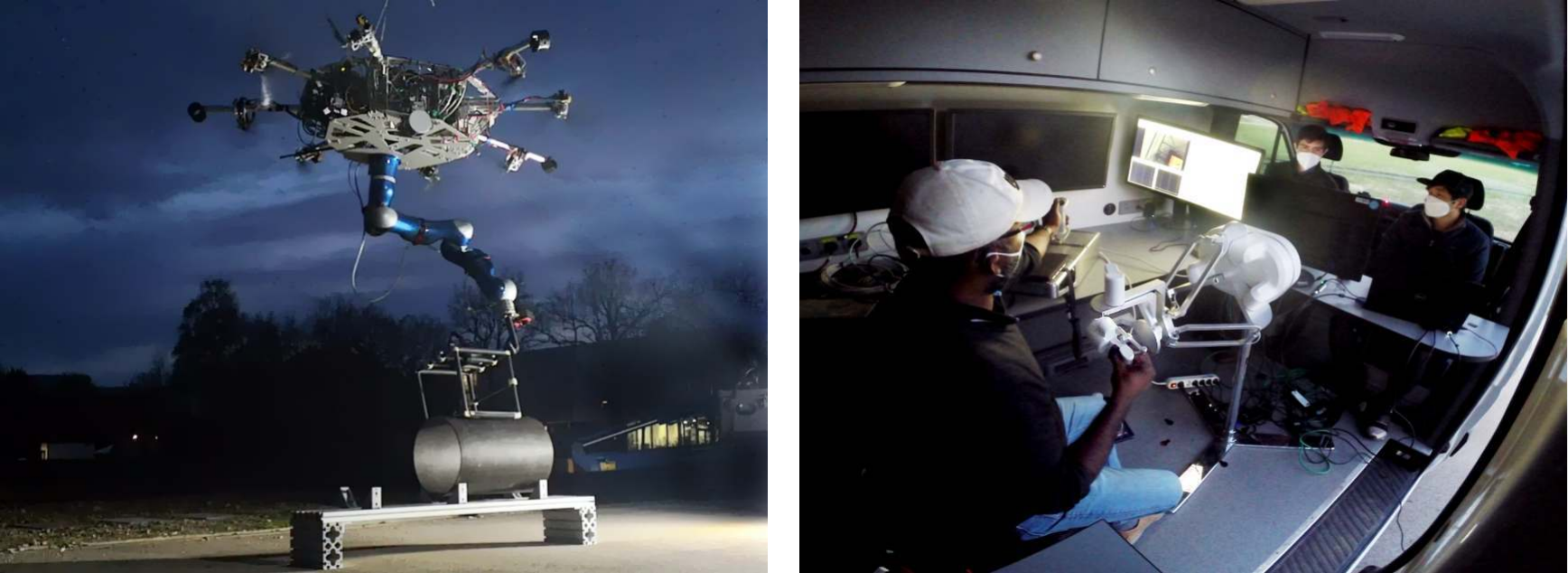}
 \caption{Left: the cable-Suspended Aerial Manipulator, dubbed SAM \citep{sam2019} during field experiment. Right: a ground station where an operator remotely controls the robotic arm through a haptic interface. In real world applications of bilateral teleoperation, the operator is often remotely located without visual contact to the robot.}
  \label{fig:introduction:1}
\end{figure}

\begin{figure}
  \centering
  \includegraphics[width=1\textwidth]{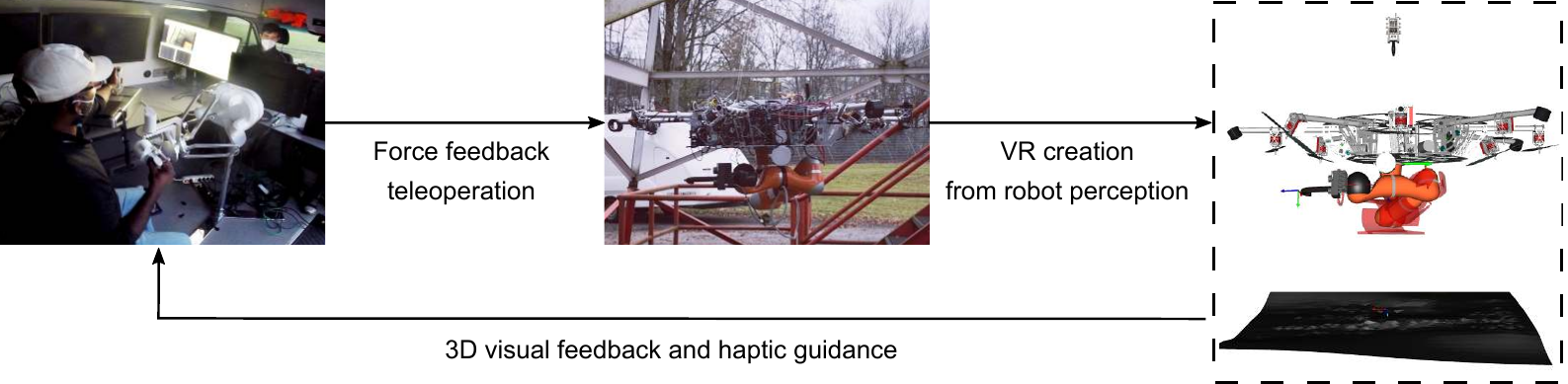}
 \caption{The proposed \green{telepresence system} with VR from robot perception and active learning. \red{In the proposed system, the robot creates VR of its workspaces as a 3D visual feedback to the human operator, and further provides a haptic guidance. The main novelty of this work is the realization of such a system for real world scenarios.}}
  \label{fig:introduction:2}
\end{figure}

Building upon the aforementioned developments, we propose a novel virtual reality (VR)-based \green{telepresence system} for an aerial manipulation system operating in industrial scenarios. Figures~\ref{fig:introduction:1} and \ref{fig:introduction:2} illustrate the main idea. The proposed system is intended for  real world scenarios, where the remotely located robot performs aerial manipulation tasks, while its human operator is inside a ground station without having direct visual contact with the robot (Figure~\ref{fig:introduction:1}). To this end, we propose a system which does not only involve a haptic device to enable the \textit{sense of touch} for the operator, but also a VR to increase the \textit{sense of vision} (Figure~\ref{fig:introduction:2}). \red{While the live video streams can also provide a certain level of situation awareness to the operator, several studies confirm that adding a virtual environment where one can change its sight-of-view, zoom in and out, and further provide haptic guidance, supports the operator in accomplishing the tasks \citep{deleveraging, whitney2020comparing, Huang2019}. Our own field studies also confirm that augmenting live video streams with 3D visual feedback and haptic guidance can enhance manipulation capabilities of aerial robots}.

\green{The main novelty of our VR based concept is its realization with a \textit{fully on-board perception} system for \textit{a floating-base robot}, which \textit{does not rely} on any external sensors like Vicon, or any pre-generated maps in \textit{outdoor environments}. Instead, \textit{multiple sensors}, namely LiDAR, a monocular camera, a pair of stereo cameras and inertial measurements units (IMUs) are jointly utilized (Table~\ref{table:comparison:relatedworks}).} \blue{To achieve this, we propose object pose estimation and active learning pipelines. First, in order to virtually display industrial objects with known geometry, we provide a simple extension of a marker tracking algorithm~\citep{artoolkit} by combining with on-board Simultaneous Localization And Mapping (SLAM). Second, if the objects of interests are geometrically unknown, we devise a LiDAR based pose estimation pipeline that combines LiDAR Odometry And Mapping (LOAM \citet{zhang2014loam}) with a pose graph, a point cloud registration algorithm \citep{besl1992method}, and a Deep Neural Network (DNN) based object detector \citep{lin2017focal}. For both the cases, the combinations are facilitated by an introspection~\citep{grimmett15introspective}  module that identifies the reliability of the pose estimation. Finally, we present a pool based active learning pipeline, which uses an explicit representation of DNN's uncertainty, to generate the most informative samples for a DNN to learn from. This enhances the sample efficiency of deploying DNN based algorithms in outdoor environments. We identify certain real world challenges and describe in detail how these introspective approaches can mitigate these challenges.}

\red{With the DLR's SAM platform \citep{sam2019}, the feasibility and benefits of the proposed idea are examined.} To this end, we first present ablation studies on the designed pipelines with indoor and outdoor data-sets from the robot sensors. Here, the influence of each component is examined with regard to mitigating the identified challenges, and we show the feasibility of creating the real-time VR, which can closely match the real workspaces of the robot. Moreover, the effectiveness of the proposed method is shown through outdoor experiments within the considered industrial scenario. This scenario, which was designed under the scope of EU project AEROARMS \citep{aeroarms}, is relevant to inspection and maintenance applications for gas and oil industry. It involves pick-and-place and force-exertion tasks during the mission, which is to deploy a robotic crawler for automating pipe inspection routines. Moreover, the SAM platform executing peg-in-hole tasks with a margin of error less than 2.5 mm is further considered, which is one of the standard manipulation tasks in industrial settings. \red{By executing over 70 executions of the aforementioned tasks over days and nights, from spring to winter, and with different users and locations, the benefits of our VR based \green{telepresence} concept are illustrated for enhancing aerial manipulation capabilities in real world industrial applications.}

In summary, the key contributions of this work are:

\begin{itemize}[noitemsep,nolistsep]
\item We propose an advanced VR based \green{telepresence system} for aerial manipulation, which provides a 3D visual feedback and a haptic guidance. The system neither requires any external sensors nor pre-generated maps, has been evaluated outside laboratory settings, and can cope with the challenges of a floating-base system. \green{Moreover, multiple sensors are fused to exploit their respective strengths for the given perception tasks.}
\item We devise object pose estimation and active learning pipelines to realize the proposed system in dynamic and unstructured environments. \blue{Challenges to existing methods are reported, and several ablation studies are provided to validate the proposed approaches. Methodologically, this work suggests the relevance of robotic introspection in realizing VR based telepresence robots with aerial manipulation capabilities.}
\item We perform exhaustive flight experiments over extended durations including 40 task executions in outdoor environments, 27 task executions within a user validation study, and the operation of the system at night. Thus, we establish the proposed concept as a viable future option for real world industrial applications.
\end{itemize}

\red{The paper starts with a survey of related work (Section \ref{sec:relatedworks}) and provide the system description of SAM robot hardware, human-machine interfaces, sensor choices, and integration (Section \ref{sec:system}).} We formulate the problem of the VR creation, and identify challenges \red{in realizing the system} (Section \ref{sec:problem}). \red{Then, the designed pipelines are presented, which are to address} these challenges (Section \ref{sec:method1}). In Section \ref{sec:evaluation}, we provide ablation studies to validate the designed framework, while Section \ref{sec:experiments} contains the results of our flight experiments. We report the lessons learned in Section \ref{sec:lessons} and conclude the work with some future extensions in Section \ref{sec:conclusion}.

\textbf{Relation to Previous Publications}\red{ This paper extends the author's previous publications}, namely \citet{lee2020visual} and \citet{lee2020estimating}. In terms of methodology, we provide a LiDAR based pose estimation pipeline (Section \ref{subsec:unknownobject}). This extension enables the creation of VR without relying on markers, which is required in industrial scenarios. The devised active learning pipeline for object detection (Section \ref{sec:method2}) extends and brings the previous theoretical framework \citep{lee2020estimating} to practical applications. Furthermore, with respect to experimental contributions, this article provides new ablation studies that are associated with the new methods. Most importantly, exhaustive outdoor experiments for  manipulation tasks are further performed to examine the benefits of the proposed VR based concept over extended durations and characterize its technical readiness for industrial applications.

\section{Related Work}
\label{sec:relatedworks}

The proposed VR based concept advances the area of VR interfaces for robotics. The comparison of \green{this work} to existing works is summarized in Table~\ref{table:comparison:relatedworks}. \red{The current literature from different domains of robotic research is discussed, which are}, pose estimation (\sectionref{subsec:knownobject} and \ref{subsec:unknownobject}), and active learning with DNNs (\sectionref{sec:method2}). \red{Importantly, we stress that this work is not to advance the state-of-the-art methods in these two areas. Rather, the aim is to apply and extend them to realize a working system for the given industrial scenarios. For example, the provided extension of a marker tracking algorithm with visual-inertial SLAM is not the main contribution of this paper.} Lastly, we further locate our work within the literature of aerial robotic perception in field applications.

\begin{table*}
\centering
\caption{\green{Comparisons between the existing VR based robotic systems and the proposed system.}}
\begin{tabular}[ht]{ccccc}
\toprule
\multicolumn{1}{c}{} & \multicolumn{1}{c}{\textbf{Outside}} & \multicolumn{1}{c}{\textbf{\textbf{No external}}} & \multicolumn{1}{c}{\textbf{Floating-base}} & \multicolumn{1}{c}{\textbf{Multiple}} \\
\multicolumn{1}{c}{} & \multicolumn{1}{c}{\textbf{the laboratory}} & \multicolumn{1}{c}{\textbf{\textbf{sensors or pre}}} & \multicolumn{1}{c}{\textbf{manipulation}} & \multicolumn{1}{c}{\textbf{exteroceptive}} \\
\multicolumn{1}{c}{} & \multicolumn{1}{c}{\textbf{settings?}} & \multicolumn{1}{c}{\textbf{\textbf{generated map?}}} & \multicolumn{1}{c}{\textbf{system?}} & \multicolumn{1}{c}{\textbf{sensors?}} \\
\midrule
AeroVR \citep{yashin2019aerovr} & \xmark  & \xmark  & \cmark &  \xmark \\
ARMAR-6 \citep{pohl2020affordance} & \xmark   & \cmark  & \xmark &  \xmark \\
ModelSegmentation\citep{kohn2018towards} & \xmark   & \cmark & \xmark & \xmark\\
AvatarDrone \citep{kim2021toward} & \xmark   & \xmark  & \cmark & \xmark \\
PaintCopter \citep{Vempati2019} & \xmark   & \xmark  & \cmark & \xmark \\
AR \citep{liu2020augmented}  & \xmark & \cmark & \xmark & \xmark \\
AR \citep{puljiz2020hololens} & \xmark   & \cmark  &  \xmark & \xmark \\
GraspLook \citep{ponomareva2021grasplook} & \xmark   & \cmark  &  \xmark & \xmark \\
The proposed system & \cmark   & \cmark  & \cmark & \cmark\\
\bottomrule
\end{tabular}
\label{table:comparison:relatedworks}
\end{table*}

\textbf{Virtual Reality Interfaces} In the past, several VR interfaces have been widely utilized in robotics including aerial systems \citep{wonsick2020systematic}. So far, the presented approaches often create the VR either by using external sensors such as Vicon and a-priori generated maps. Notably, \citet{Vempati2019} utilizes a-priori generated maps for the applications of VR in aerial painting. For aerial manipulation, \citet{yashin2019aerovr} uses Vicon system to create the VR while \citet{kim2021toward} renders the environment with a portable sensor kit \citep{oh2017technical}. Recently, many VR techniques have gained interest in the robotic manipulation community. Therein, many works \citep{haidu2020automated, zhang2020graph} let a human perform demonstration in VR, and transfer the demonstrated manipulation skills to real robots. These works greatly show the synergy between VR and robotics. \red{As this paper demonstrates the feasibility of creating VR with on-board sensors only, the work can contribute to many of these works in showing how one can create a VR for robotics. }

On the contrary, many researchers aimed to provide  VR of the remote scene by applying 3D reconstruction techniques \citep{ni_song_xu_li_zhu_zeng_2017, kohn2018towards}. For example, \citet{kohn2018towards} presents an approach using RGB-D camera. As the main challenge of reconstruction based methods is the limited bandwidth in communication, \citet{kohn2018towards} proposes an object recognition pipeline, i.e., replace the detected object with sparse virtual meshes and discard the dense sensor data.  \citet{pohl2020affordance} uses RGB-D sensor to construct a VR for affordance based manipulation with a humanoid, while \citet{liu2020augmented} and \cite{puljiz2020hololens}  create augmented reality for a drone and a manipulator, respectively. \citet{deleveraging} conducts a user study, and argues that the point clouds of RGB-D sensors are noisy and inaccurate (with artifacts), which motivates for point cloud pre-processing methods for \green{telepresence applications}~\citep{deleveraging}. In contrast, our approach is based on scene graphs (Section \ref{sec:problem}) with pose estimation, which is an alternative to 3D reconstruction methods. \green{Finally, the main novelties are illustrated in Table~\ref{table:comparison:relatedworks}, which are the realizations of a VR based telepresence system for outdoor environments using multiple sensors jointly. No external sensors or pre-generated maps are used, while dealing with specific challenges of a floating-base manipulation system, i.e., the surface that holds a robotic arm is constantly changing over time, thereby inducing motions for the attached sensors.}

\textbf{Object Pose Estimation} One of the crucial components in the proposed framework is object pose estimation algorithms. This is because we utilize a scene graph representation, which requires 6D pose of the objects for creating a 3D display, as opposed to a 3D reconstruction of the remote site. As the literature is vast, we refer to the survey \citep{he20216d} for a comprehensive review. \red{In this work, the main novelty is the working solutions for the considered application, which is tailored towards realizing the proposed VR system. For this, the two scenarios are discussed below. These are visual object pose estimation for objects of known geometry, and LiDAR based method for unknown geometry.}
 
If the object is known and accessible a-priori, one of the robust solutions is to use fidicual marker systems. Fidicual markers, which create artificial features on the scene for pose estimation, are widely used in robotics. The use-cases are for creating the ground truths \citep{Wang2016},  where environments are known \citep{Malyuta2019}, for simplifying the problem in lieu of sophisticated perception \citep{Laiacker2016}, and also calibration and mapping \citep{nissler18simultaneous}. \red{However, as the herein aim is on real-time VR creation, this use-case demands stringent requirements on their limitations in run-time, inherent time-delays and robustness. Therefore, an extension of ARToolKitPlus is provided \citep{artoolkit} with an on-board visual-inertial SLAM system. }

For LiDAR, point cloud registration is often used for pose estimation. By finding the transformation between the current scans and a CAD model of an object, we can obtain 6D pose of an object. Broadly, point cloud registration algorithms can be classified as local \citep{park2017colored, rusinkiewicz2001efficient, besl1992method} or global \citep{zhou2016fast}, and model based \citep{pomerleau2015review} or learning based \citep{wang2019deep, zhang2020deep}. \red{As CAD models of objects are often not available in the given industrial scenario, a DNN based detector and the idea of LOAM with pose graphs are combined, in order to obtain robust object pose estimates that cope with occlusions, moving parts and view point variations in the scene.}

\textbf{Active Learning for Neural Networks} \red{The motivations are the considerations of  field robotic applications of DNN based object detectors.} Here, the need for labeled data can cause overhead in development processes, especially while considering a long-term deployment of learning systems in outdoor environments. For example, weather conditions can change depending on seasons, and we need to efficiently create labeled data. Active learning provides a principled way to reduce manual annotations by explicitly picking data that are worth being labeled. One way to autonomously generate the "worth" of an unlabeled sample is to use uncertainty of DNNs. In the past, for robot perception, we find active learning frameworks using random forests, Gaussian processes, etc \citep{narr16stream, mund15active} while for DNNs, \citet{mackay1992information} pioneered an active learning approach based on Bayesian Neural Networks, i.e., a probabilistic or stochastic DNN \citep{gawlikowski2021survey}, which offers a principled method for uncertainty quantification. Recent works can also be found on active learning for DNN based object detectors \citep{choi2021active, aghdam2019active}, where the focus is on adaptations of active learning to existing object detection frameworks. These include new acquisition functions (or selection criteria) and how uncertainty estimates are generated.

For uncertainty quantification in DNNs, so-called Monte-Carlo dropout (MC-dropout \citet{gal2016dropout}) has gained popularity recently. The main advantage of MC-dropout is that it is relatively easy to use and scale to large data-set. However, MC-dropout requires a specific stochastic regularization called dropout \citep{srivastava2014dropout}. This limits its use on already well trained architectures, because the current DNN based object detectors are often trained with other regularization techniques such as batch normalization \citep{ioffe2015batch}. Deep ensemble \citep{lakshminarayanan2016simple} is another scalable framework with a relaxed assumption on the model. Unfortunately, deep ensemble requires training of several large DNN models to form an ensemble. This technique is popular generally, but it is difficult to be utilized in active learning due to the inefficiency in training. \red{In this article, a previous work \citep{lee2020estimating} on uncertainty quantification of DNNs is instead utilized.} The main motivations are the scalability to large architectures and data-sets, training-free feature that needs no changes in network architectures and no re-training, and the ability to model every layer of DNNs as Bayesian. These aspects can make the given framework well suited for active learning in practice, and \red{thus, this work attempts to provide an extension to active learning for its real world applications in robotics.}

\textbf{Aerial Robotics: Perception in Outdoor Environments} \red{The research area on the aerial robotic perception in outdoor environments is a fast growing field with several ground breaking results. For example, \citet{saska2017system, saska2014swarms} pioneered the area of swarm robotics, while \citet{loquercio2021learning, foehn2022agilicious} demonstrated impressive results in agile flights of micro aerial vehicles. Aerial robotics, with fully on-board perception, have also been part of the recent DARPA subterranean challenges~\citep{rouvcek2021system, tranzatto2022cerberus, Agha2021NeBulaQF, hudson2021heterogeneous}. Vision based localization methods have also made tremendous progress \citep{ebadi2022present, weiss2012real, scaramuzza2014vision, lutz2020ardea}. We note that, on the other hand, this paper contributes to orthogonal areas namely, VR, telepresence robots and aerial manipulation, which differs from tackling navigation problems for aerial robots.}

\section{System Description, Problem Statement and Identified Challenges}
\label{sec:system:and:problem}

\green{This paper investigates} how a robot can create a VR of a remote scene using on-board sensors and computations. This is to enhance the situational awareness of the human operator in real world applications. To set the scene for the work, we first describe the system integration that are needed to implement \red{the proposed VR based telepresence concept}. \red{Then, the problem of VR creation, using on-board sensors with a scene graph approach, is formulated.} The limitations of the off-the-shelve methods are then presented, which hinder realization of the proposed system in outdoor environments. 

\subsection{System Description}
\label{sec:system}

\begin{figure}
  \centering
  \includegraphics[width=1.0\textwidth]{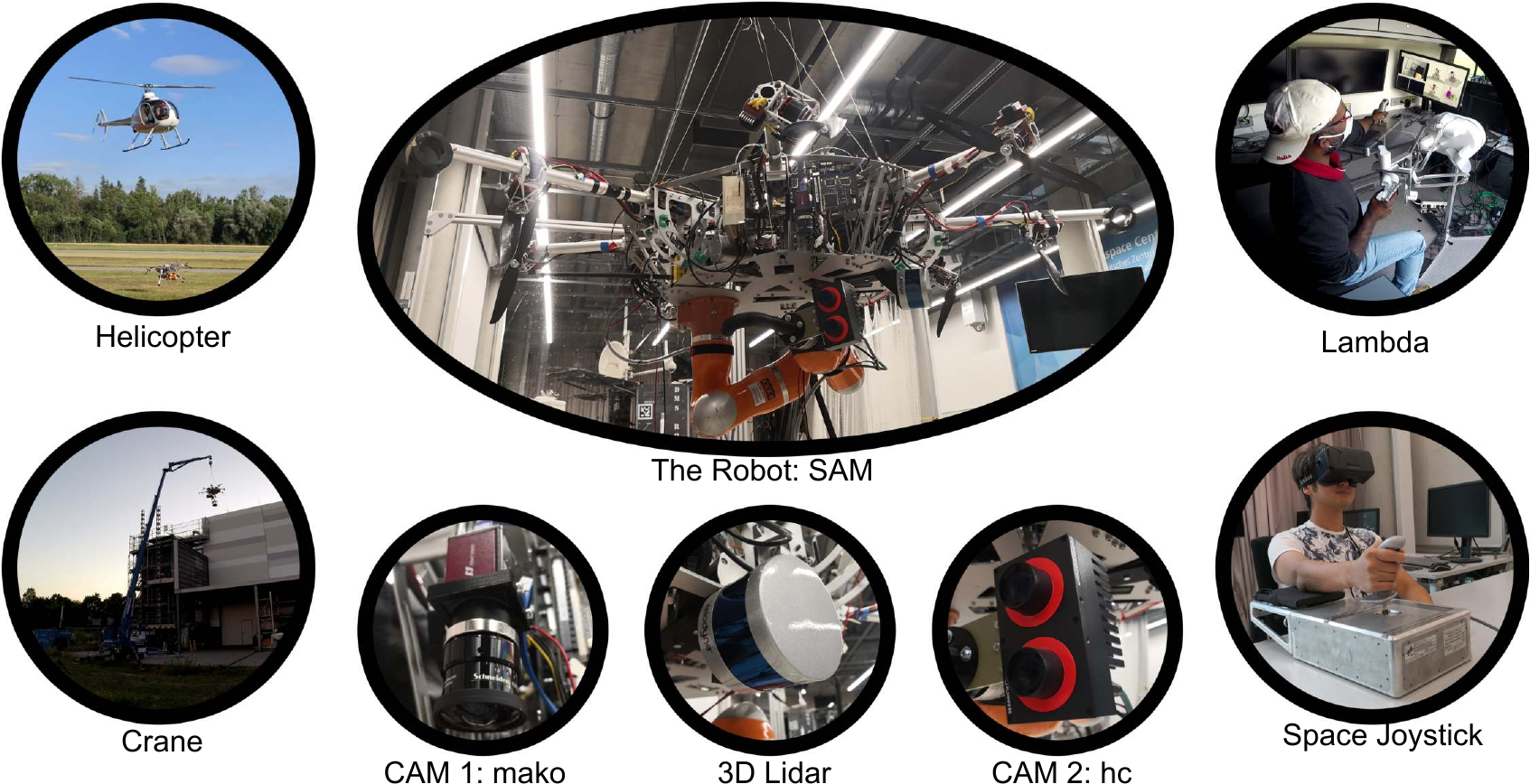}
 \caption{The concept of SAM with its integrated sensors and human-machine interfaces. Left: the concept involves the carriers such as a manned helicopter or a crane, which transports SAM to a desired location. Middle: SAM is equipped with a stereo camera at the end effector or the manipulator, a monocular camera as well as a LiDAR on the cable suspended platform. Right: haptic interfaces are integrated for teleoperating the robotic arm. Robot hardware constitutes of SAM, and the carriers (helicopter or crane), while the used haptic devices are Force Dimension Lambda and Space Joystick. Finally, the integrated sensors are a monocular camera (dubbed mako), a 3D LiDAR, and a stereo eye-in-hand camera (dubbed hc).}
  \label{fig:system:1}
\end{figure}

\begin{figure}
  \centering
  \includegraphics[width=1\textwidth]{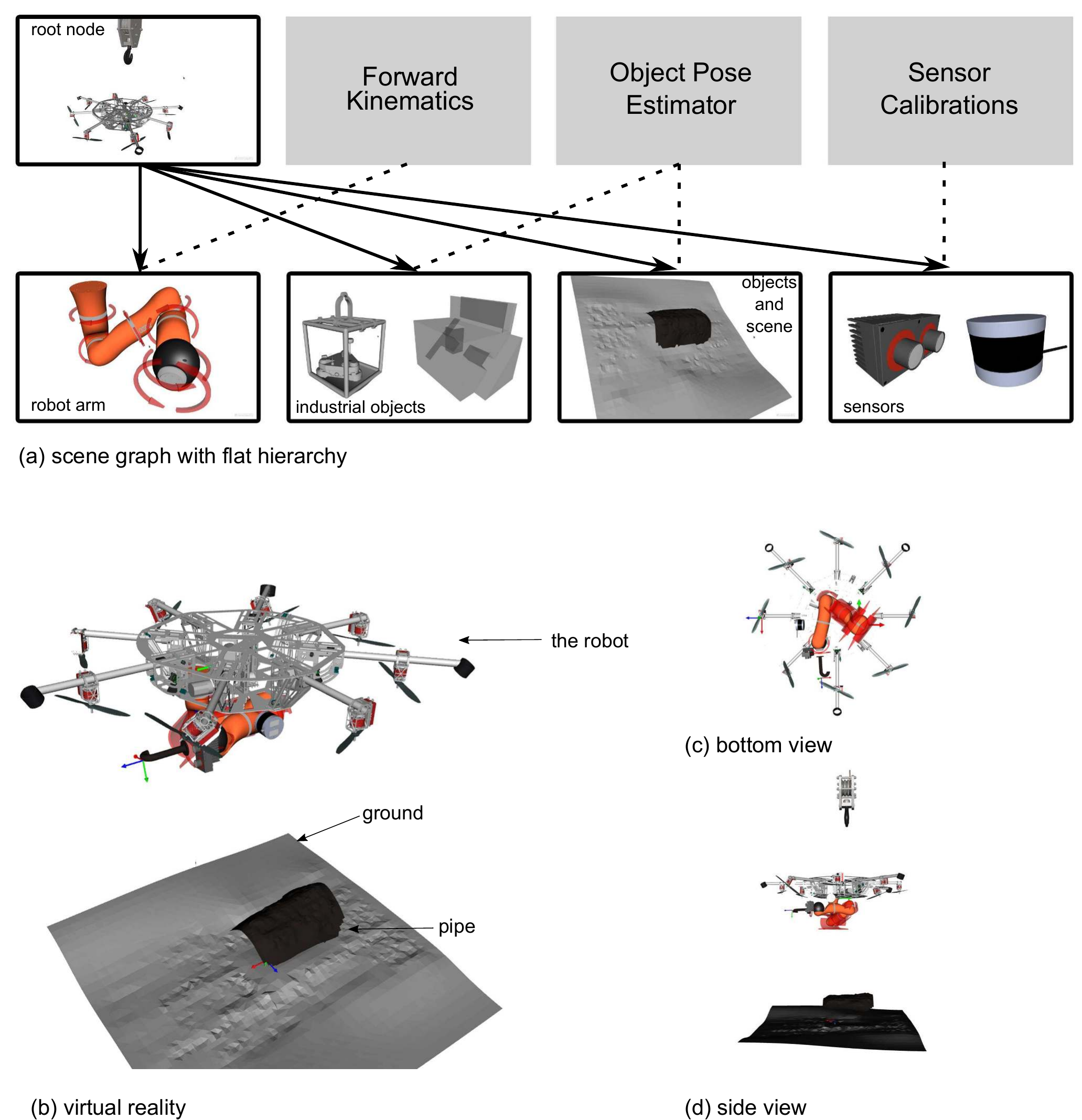}
 \caption{Illustration of the scene graph representation for the proposed VR framework. The root node is the base frame of the robot, while robot arm, industrial objects, scenes, and sensors are object nodes with transformation matrices as the edges. Forward kinematic provides state of the robot arm, and the fixed transformations to the robot sensors are obtained from extrinsic sensor calibrations. Then, the 6D estimates of object pose are obtained online using robot perception. \red{Therefore, this work focuses on the object pose estimation for the realization of the proposed system.} }
  \label{fig:problem:1}
\end{figure}

This section describes the used robotic systems with a focus on \green{robot hardware, haptic device and VR interfaces, and sensors. Main features of the system is also discussed.} Figure~\ref{fig:system:1} depicts an overview of our physical hardware.

\textbf{Robot Hardware} DLRs' SAM \citep{sam2019} is a novel aerial manipulation system for inspection and maintenance applications. SAM is composed of three modules, namely a carrier, a cable suspended platform and a seven degrees of freedom (DoF) industrial robotic arm - KUKA LWR \citep{kuka}. The purpose of the carrier is to transport the manipulation system to a desired location. We use a crane in this work which provides safety, versatility, robustness and applicability for \green{the} considered industrial scenario\footnote{\red{On the other hand, there is no free lunch. Cranes may not be able to reach all the desired location as they require available access routes by ground. There are also several industrial tasks where smaller robotic arms with less DoF may be sufficient.}}. Then, a platform attached to the carrier via a rope, autonomously damps out the disturbances induced by the carrier, the environment, and the manipulator. This oscillation damping control is performed using eight propellers and three winches. \green{Another important component of our system is the seven Dof torque controlled KUKA LWR \citep{kuka}, which features significantly more powerful versatile manipulation capabilities than many existing smaller manipulators. The main feature of the cable-suspension concept is that the weight of SAM is supported by the carrier. Thus, the required  energy to carry an aerial robot arm can be reduced. This allowed us to scale down the overall size, from a helicopter based system \citep{Kondak2014} to a relatively smaller robot, which enables operation in confined spaces. The helicopter based system had two rotors with overall diameter of 3.7m, while SAM can fit within 1.5m diameter. Moreover, the cables from the carrier can also be used to power SAM, which gives theoretically unlimited operation time. In the appendix, more details are provided regarding the platform control and different architectures.} \citet{sam2019} \green{can also be referenced for conceiving} design and control aspects of SAM in detail.

\textbf{Haptic Devices and Virtual Reality Interfaces} \red{In this work, two haptic devices, namely a space qualified haptic device called the Space Joystick RJo \citep{spacejoystick}, and also a six Dof force feedback device, Lambda (Force Dimension), are integrated in order to teleoperate the LWR on SAM. This work's VR interface is based on Instant Player \citep{hulin2012}, which is a lightweight software that runs on standard laptops without GPUs (enhancing portability). Instant Player also supports various hierarchies of a scene graph to create the required display. Facebooks' head mounted display Oculus is also integrated as an option and use Ubiquiti Bullet for the WiFI connection. The robot is equipped with advanced control strategies, namely whole body teleoperation, and adaptive shared control. The time-domain passivity approach of \citet{spacejoystick} is employed to obtain stable teleoperation control under communication time delays, packet loss and jitters.} These control methods advance aerial manipulation capabilities.  \citet{9476739} and \citet{balachandran2021finite} present these concepts in more detail. The former presents a passivity based framework to enable time-delayed teleoperation of different hierarchically-sorted tasks through the use of multiple input devices.
\citet{balachandran2021finite} present a  method to stabilize on-line adaptation of control authorities for the operator and the virtual assistance system in haptic shared control.

\textbf{Sensor Choices and Integration} We integrate \red{several sensors} for measuring the robot's own states as well as to perceive the environment. \red{More specifically, a KUKA LWR \citep{kuka} is equipped with torque and position sensors, which measure its joint torques and angles. Furthermore, we integrate other sensors on SAM for the perception tasks.} Firstly, a camera (the Allied Vision: mako) is integrated on the frame of SAM to stream the overall operational space of the robotic arm. This is because the operator prefers an eye-to-hand view, which is more natural to a human. The camera provides color images of 1292 by 964 px at 30Hz. Secondly, a stereo camera is integrated near the tool-center-point (tcp) of the robotic arm. This eye-in-hand set-up avoids occlusion of the camera view by the robotic arm, and ensures proximity to the considered objects. These are crucial for the success of our image processing algorithms\red{, i.e., the accuracy of visual marker tracking depends on the size of the markers and their distance to the sensors, while the depth sensing from the stereo depends on the baseline.} We use a commercial 3D vision sensor \red{the Roboception} \red{Rcvisard that provides built-in visual-inertial SLAM. The SLAM system originates from \citep{schmid2014autonomous, lutz2020ardea} but we refer the reader to the company for more details.} Rcvisard streams 1280 by 960 px images at 25Hz and SLAM estimates can be acquired at 200Hz by fusing it with an IMU. Lastly, as a step towards industrial application of SAM, we mount Velodyne PUK-LITE LiDAR on the frame of SAM, which provides 3D point clouds of the scene at 10Hz. We intend to use LiDAR for 3D object pose estimation as well as navigation of SAM in outdoor environments. \red{Note that the minimum range is set to 0.9m while the maximum range of 100m is utilized. We designed and integrated the sensor stacks so that the close range perception is not affected.} All the perception algorithms are executed on the NVIDIA Jetson TX2.

\red{In Appendix~\ref{appendixA}, more details on platform control, telepresence systems and IT architectures are presented.}


\subsection{Problem Formulation and Identified Challenges}
\label{sec:problem}

\begin{figure}
  \centering
  \includegraphics[width=1\textwidth]{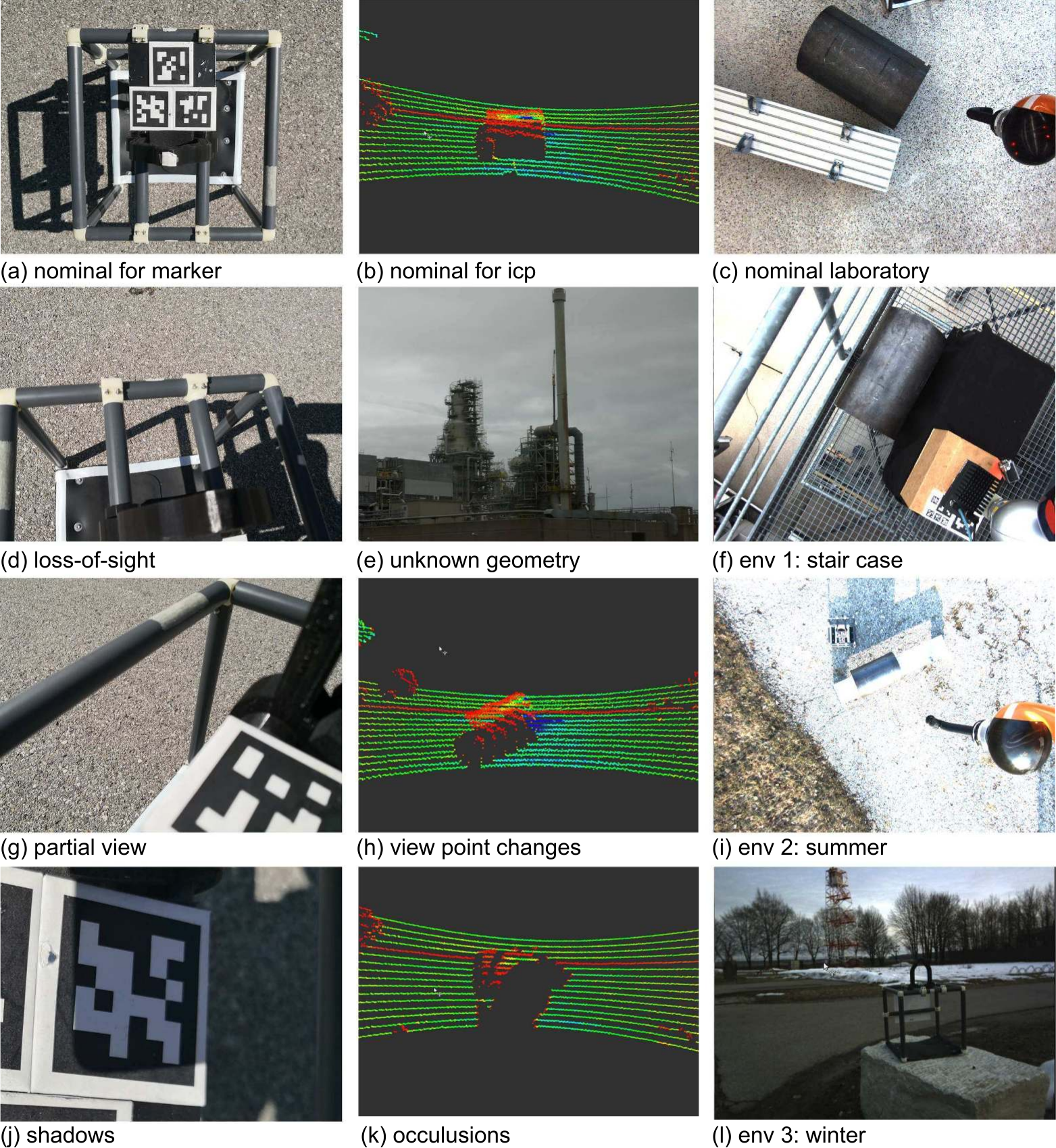}
  \caption{\red{Identified challenges for realizing our VR based concept. Top: (a,d,g,j) show the challenges associated with a marker tracking algorithm. Middle: (b,e,h,k) depict the challenges associated with directly applying point cloud registration methods for pose estimation. In particular for (e), precise geometry of objects are not available for pipe inspection scenario as an example, and therefore, its CAD models must be reconstructed online. Bottom: (c,f,i,l) visualize different scenes that a learning based method must cope with, when deployed for real world applications. For example, a DNN trained in a laboratory, may not generalize to the scenes with (f) stair cases.} }
  \label{fig:problem:2}
\end{figure}

\blue{Assume that SAM is performing manipulation tasks far away from the human operator. So, the operator does not have direct visual contact to the scene, and the robot has to enhance the situational awareness of the operator. For this, SAM creates a VR of its environment and workspaces using on-board sensing and computations, and further provides haptic guidance via virtual fixtures \citep{rosenberg1993virtual}.} \red{Followed by the system level requirements, the problem formulation and the challenges of realizing such VR based telepresence concept are introduced next (see Figures \ref{fig:problem:1} and \ref{fig:problem:2}).}

\red{The system level requirements are highlighted as follows. Firstly,} the created VR has to accurately match the real remote site in real-time. This is because the operator needs visual feedback that reflects reality, and the performance of haptic guidance depends on the positioning accuracy and run-time. The latter is due to potential movements of the robot while hovering. Second, the robustness of the created VR is crucial to give a sense of trust to the human operator and further provide reliable haptic guidance. This means that the abnormalities in the object pose estimators are to be coped with, which often arises in \red{outdoor} environments. Last, the algorithms must run on-board the robot, and \red{only send the transformation matrices through WiFi network (apart from an initialization phase, where surface reconstructed 3D models are sent).} This is to avoid overloading of the communication channel for stable bilateral teleoperation. For example, both the sparse LiDAR point clouds and the dense stereo point clouds must be processed first, and only the pose of the objects must be sent through the WiFi network. \red{The pose information contains only six float values, while continuous streaming of the point clouds require much more memory that grows with the number of points.}

\blue{For VR creation that addresses aforementioned requirements, this work relies on a scene graph approach (shown in Figure~\ref{fig:problem:1}). A scene graph is general data structure with graph or tree like representations. It is used by the VR/AR softwares \citep{hulin2012}, in order to produce the real-time 3D visualizations. Mathematically, let S be a scene graph. It constitutes of sets of nodes and edges, denoted by (V, E). The nodes V are any 3D models, while the edges E represent the spatial relationships. The root node $\text{V}_\text{root}$ is chosen to be the robot's base frame, which is a fixed coordinate of the SAM platform. Then, a flat hierarchy of the scene graph \citep{hulin2012} is assumed. This means the root node is a single "parent" to all other "child" nodes. In the given scenarios, the models to be displayed in VR are the sets of industrial objects, reconstructed external scenes, robotic arm, and the robot sensors. For a node of robotic arm $\text{V}_\text{LBR}$, the corresponding edges $\text{E}_{\text{LBR}}^{\text{root}}$ are readily provided by the forward kinematics. Similarly, the edge of three sensors, $\text{E}_{\text{hc}}^{\text{root}}$, $\text{E}_{\text{mako}}^{\text{root}}$ and $\text{E}_{\text{LiDAR}}^{\text{root}}$, are the outputs of the extrinsic camera calibration. These spatial relations or the relevant transformation matrices are therefore fixed for the sensor nodes $\text{V}_{\text{hc}}^{\text{root}}$, $\text{V}_{\text{mako}}^{\text{root}}$ and $\text{V}_{\text{LiDAR}}^{\text{root}}$.}

\blue{On the contrary, the spatial relations of industrial objects and external scenes are constantly changing, leading to the problem of pose estimation. Here, we divide the problem formulation into two sub-problems. The first sub-problem is when the object is known a-priori with available 3D models ($\text{V}_{\text{o1}}^{\text{root}}$, $\text{E}_{\text{o1}}^{\text{root}}$), while the second sub-problem is when the object is semantically known a-priori, but no primitives on the geometry exist ($\text{V}_{\text{o2}}^{\text{root}}$, $\text{E}_{\text{o2}}^{\text{root}}$). For the former, the corresponding edges are to be estimated. The latter involves the estimation of both the nodes and the edges.} As articulated in Section \ref{sec:system}, \blue{the available} raw sensor data are RGB camera images $\mI \in \mathbb{R}^{H \times W \times 3}$, where $H$ and $W$ are the image height and width, respectively. The images are obtained either from the eye-in-hand stereo camera (denoted by hc), or a monocular camera at the base (denoted by mako). A LiDAR, which is located also at the base, generates scans that are represented by the point clouds $\mP = (\vp_1, \vp_2, ..., \vp_N) \in \mathbb{R}^{3 \times N}$. We also have a visual-inertial SLAM system at the end effector of the robotic arm, which outputs the rotation matrix $\mR$ and translation vector $\vt$ between the coordinate frames of the camera and a fixed world frame. \blue{In summary, the problem of VR creation can be formulated as estimating $\text{E}_{\text{o1}}^{\text{root}}$, $\text{V}_{\text{o2}}^{\text{root}}$ and $\text{E}_{\text{o2}}^{\text{root}}$ using the available sensory data from different cameras, an IMU and a LiDAR.}

\blue{For this problem, several existing approaches can be applied. However, several practical challenges of directly applying these approaches have been identified from the field work (depicted in Figure~\ref{fig:problem:2}).} For objects of known geometry, we can resort to marker based object pose estimation methods. For this, we cannot assume the holistic view of the markers. Violations of this assumption are caused by shadows, loss-of-sight or partial views of the markers. This results in failures while using off-the-shelf marker tracking methods. Moreover, in an industrial scenario, we cannot assume the availability of precise CAD models. Thus, point cloud registration methods cannot be directly employed. Tracking is also subject to occlusions and moving objects in front of the LiDAR, e.g. the robotic arm, and significant view point changes also result in less accurate 6D object poses while employing off-the-shelf methods such as iterative closest point algorithm. Finally, while deploying data-driven approaches for field robotics, key to its success is preparation of the data. The main challenges are the variations of scenes encountered during long-term deployment; \red{darkness in the evenings or snow in winter are such examples.} This means data has to be repeatedly collected for varying environment conditions, which is a laborious process. So, the question is: how to make the data collection procedure more efficient so that DNNs can generalize. \blue{In the next section, these challenges are revisited after providing mathematical formulations of existing methods, which is then followed by our extensions to resolve these challenges.}

\section{The Proposed Methods}
\label{sec:method1}

\red{The aim} is to create a 3D display of the robot and the objects so that the human operator can remote control the robotic arm from a distance. If done in real-time, the operator can \textit{see} the VR and perform the tasks. Haptic guidance via virtual fixtures can further help the human operator during the execution of challenging manipulation tasks. So far, we have formulated the problem and also outlined the practical challenges. As previously discussed, the scene graph creation problem relies on the accurate, fast and reliable 6D object pose estimation algorithms for the industrial objects of known and unknown geometries. This section describes the proposed pipeline for the object pose estimation. 

\subsection{The Proposed Pipeline for Objects of Known Geometry}
\label{subsec:knownobject}

\begin{algorithm}
   \caption{Robust marker localization algorithm with Visual-Inertial SLAM}
   \label{alg:method1:1}
   \SetAlgoLined
   \SetKwInOut{Input}{input}
   \SetKwInOut{Output}{output}
   
   \Input{
        \begin{tabular}{l l}
        $\mI$ & camera images from the eye-in-hand (hc) camera. \\
        $m$   & target marker identification. \\
        $i$   & identification numbers of additional markers $i=1,2,..n$. \\
        $t_d$ & time delay parameter, either online computed or prespecified. \\
        $\mT^{w}_{{hc}}$ & SLAM estimates of hc camera w.r.t a world coordinate. \\
        \end{tabular}
   }
   \Output{
        \begin{tabular}{l l}
        $\mT^{{hc}}_{m}$ & 6D pose of the target marker m w.r.t the hc camera. 
        \end{tabular}
   }
   \BlankLine
   \Begin{
   \tcc{Initialization}
   $\mT_{m}^{{hc}}$(0), $\mT_{i}^{{hc}}$(0) $\leftarrow$ multiART+($\mI$)  $\forall i$ \tcp*{detect all the markers (Equation \ref{eq:method1:homography})}
   $\mT_{m}^{i}$ $\leftarrow$ marker$\_$init($\mT_{m}^{{hc}}$(0),$\mT_{i}^{{hc}}$(0)) $\forall i$ \tcp*{save all the relative poses}
   \BlankLine
   \tcc{Main Loop}
   \While(){\emph{True}}
   {
      $\mT_{m}^{{hc}}$(t), $\mT_{i}^{{hc}}$(t) $\leftarrow$ multiART+($\mI$)  $\forall i$ \tcp*{detect the markers (Equation \ref{eq:method1:homography})}
      \uIf{\emph{all markers detected}}{
      $\mT_{m}^{{hc}}$(t), $\mT_{m,i}^{{hc}}$(t) $\leftarrow$ trafo2m($\mT_{m}^{{hc}}$(t), $\mT_{i}^{{hc}}$(t), $\mT_{m}^{i}$) $\forall i$ \tcp*{transform to target}
      $\mT_{m}^{{hc}}$(t) $\leftarrow$ ransac$\_$avg($\mT_{m}^{{hc}}$(t), $\mT_{m,i}^{{hc}}$(t)) $\forall i$ \tcp*{ransac and average}
      $\mT_{m}^{i}$ $\leftarrow$ init$\_$update($\mT_{m}^{{hc}}$(t), $\mT_{i}^{{hc}}$(t)) $\forall i$ \tcp*{update all the relative poses}
      }
      \uElseIf{\emph{not all marker detected}}{
$\mT_{m}^{{hc}}$(t), $\mT_{m,i}^{{hc}}$(t) $\leftarrow$ trafo2m($\mT_{m}^{{hc}}$(t), $\mT_{i}^{{hc}}$(t), $\mT_{m}^{i}$) $\forall i$\tcp*{transform to target}
      $\mT_{m}^{{hc}}$(t) $\leftarrow$ ransac$\_$avg($\mT_{m}^{{hc}}$(t), $\mT_{m,i}^{{hc}}$(t)) $\forall i$ \tcp*{ransac and average}
      }
      \uElseIf{\emph{no marker detected}}{
      $\mT_{m}^{{hc}}$(t) $\leftarrow$ slam$\_$integrate($\mT_{{w}}^{{hc}}(t)$, $\mT_{{hc}}^{{w}}(t-\Delta t)$, $\mT_{t}^{{hc}}(t-\Delta t)$) \tcp*{Equation \ref{eq:vio_integrate}}
      }
      $\mT_{m}^{{hc}}(t+t_d)$ $\leftarrow$ delay$\_$integrate($\mT_{w}^{{hc}}(t+t_d)$, $\mT_{{hc}}^{w}(t)$, $\mT_{m}^{{hc}}(t)$) \tcp*{Equation \ref{eq:vio_compensator}}
   }
   }
\end{algorithm}

\red{Once the objects to be actively manipulated are known a-priori, i.e., the CAD models are available and the objects are physically accessible, the fiducial marker systems \citep{artoolkit} can be exploited.} These systems consist of a marker, which is a physical plane with black and white squared shapes (similar to QR codes), \red{and a detection with a decoding algorithm}. The key idea is to artificially create features on a plane that are physically attached to an object. Then, we can compute the pose of a camera in relation to a coordinate of the plane via a homography. Concretely, using the eye-in-hand (hc) camera, the goal is to find the transformation matrix of the markers $\mT_{{m}}^{{hc}}$, expressed in the coordinate system of the camera, which constitutes of the rotation matrix $\mR_{{m}}^{{hc}}$ and the translation vector $\vt_{{m}}^{{hc}}$. To do so, four corner points of the markers are extracted, which are expressed in the marker coordinates $\vp_{{m}}=\left ( x_{{m}}, y_{{m}}, 0 \right )^T$ (hence $z_{{m}}=0$ and given the size), and the image plane with pixels $\vp_{{image}}=\left ( u_{{m}}, v_{{m}}, w_{{m}} \right )^T$. Then, the optimizer:
\begin{equation}
\label{eq:method1:homography}
\begin{aligned}
    & \vh(t) = \underset{\vh}{\text{arg min}} \sum_{i=1}^4 \rho(\vp_{i,{image}}(t), \mH_{{image}}^{{m}}(t) \vp_{i, {m}}(t)) \quad \text{where}\\
    & \mH_{{image}}^{{m}}(t) = \left ( \mR_{{m}}^{{hc}}(t) + \frac{\vt_{{m}}^{{hc}}(t)}{d}\vn^T \right ),
\end{aligned}
\end{equation}
is the solution to the homography problem. Here, $t$ denotes time, $\mH_{{image}}^{{m}}$ is the homography matrix with $\vh$ being its vector form, and $\rho$ is a distance based cost function. Knowing the homography matrix, the desired rotation and translations can be obtained given the parameters of the intrinsic camera calibration: $d$ and $\vn$. Typically, an algebraic formulation is used with the Direct Linear Transformation (DLT) algorithm \citep{andrew2001multiple}. We note that the fidicual marker systems are widely adopted as ground truths in the robotics community for its accuracy \citep{Wang2016}. 

\textbf{Challenges} \red{However, many existing fiducial markers systems \citep{artoolkit, Wang2016, Malyuta2019, Laiacker2016} do not address this work's application scenarios, where aerial manipulation tasks in outdoor environments are considered.} For example, shadows that are created by the robot can often destroy certain shapes of the markers and as a result, the methods would fail as the artificial visual features in the markers are occluded. Similarly, the eye-in-hand camera can lose the view on the marker as the manipulator and the base can move rapidly. Lastly, time delays that are inherent in these systems must be corrected in order to create a real-time virtual display of the scene. \red{Next, the proposed solution to these challenges are described.}

\textbf{Our Solution} To tackle these problems, we propose a robust marker localization pipeline (depicted in Algorithm \ref{alg:method1:1}) as an extension to ArtoolKitPlus \citep{artoolkit}. As an overview, the proposed pipeline utilizes multiple markers as well as the robots' SLAM system. \red{To explain, multiple markers are placed on an object, where there exist a predefined target marker ID $m$ and n additional markers with unique identifications, i.e., $i=1,2, ..., n$.} This results in total $k=n+1$ markers. At initialization, the algorithm detects all the markers, where \textit{multiART+} is the function that executes a variant of marker tracking method: ArtoolKitPlus \citep{artoolkit}). Using the eye-in-hand camera image $\mI$ (either the left or the right camera of the stereo setup), we obtain the initial 6D pose of the target marker $\mT_{m}^{{hc}}$ as well as all n additional markers $\mT_{i}^{{hc}}$ at $t=0$. Then, we save the relative poses of all n markers to the target marker m (denoted $\mT_{m}^{i}$ for $i=1,2,...,n$). This step is executed within the function \textit{marker$\_$init}. 

\red{Then, the 6D pose of the target marker $\mT^{{hc}}_{m}$ can be obtained in the main loop of Algorithm \ref{alg:method1:1}.} The first step is to execute \textit{multiART+}. Then, if all the k markers are detected, we transform the 6D pose of n additional markers to the target marker: $\mT_{m}^{{hc}} = \mT_{i}^{{hc}}\mT_{m}^{i}$ (executed with a function \textit{trafo2m}). As this results in n additional 6D poses of the target marker m, we note them as $\mT_{m,i}^{{hc}}$ for $i=1,2,...,n$. Then, RANSAC \citep{ransac} is applied to these estimates to remove the outliers, and then we perform averaging to reduce the variance (\textit{ransac$\_$avg}). Then, the relative transformations $\mT_{m}^{i}$ are updated. If at least one marker is detected, the same step is applied to estimate the target marker without updating the relative transformations $\mT_{m}^{i}$. We also note that RANSAC is skipped when less than three points are available. The described steps have two advantages. First, the accuracy and the orientation ambiguity of ArtoolKitPlus can be improved with RANSAC, and second, the algorithm is robust to loss-of-sight of the target marker, i.e., detecting only one of the markers is enough to still estimate the 6D pose of the target. Similar steps have been presented in the past with several variants \citep{Laiacker2016, Nissler, Malyuta2019}. 

\begin{figure}
  \centering
  \includegraphics[width=1.0\textwidth]{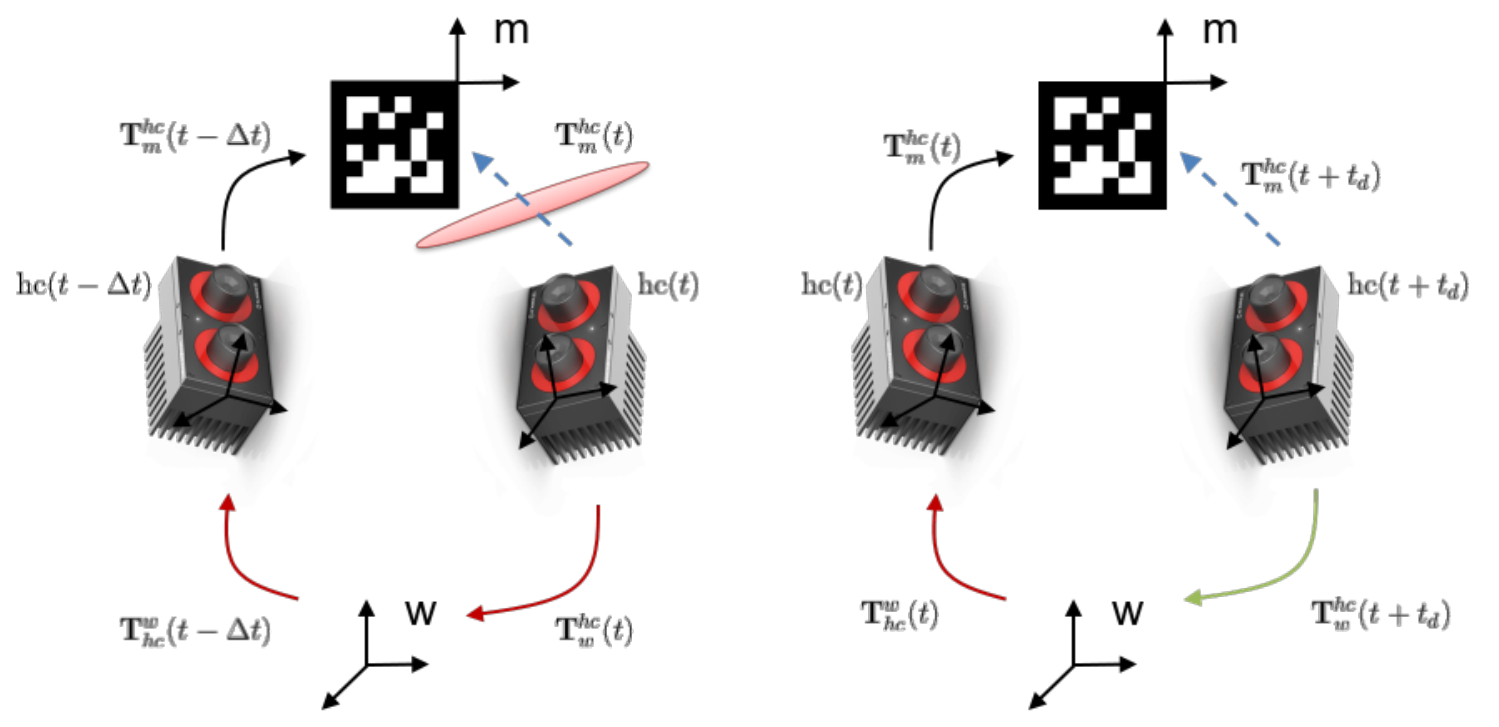}
 \caption{\red{The proposed extension of ARToolKitPlus.} Left: the position and orientation estimates of camera motion from a SLAM system infer the object when the markers are not detected. Right: \red{linear and angular velocity estimates of the SLAM system are used with the time delay term $t_d$ to predict the motion of the camera in $t+t_d$ seconds.}}
  \label{fig:method1:marker}
\end{figure}

However, the algorithm must be robust to loss-of-sight on all the markers, and further compensate for the time delay. \red{This is achieved by extending the algorithm with SLAM estimates.} The overview is depicted in Figure~\ref{fig:method1:marker}. As a first step, we propose to address the problem of complete loss-of-sight on all the markers by integrating SLAM estimates of camera motion with respect to its inertial coordinate, i.e, utilize the estimated transformation between the hc camera to a world coordinate of our SLAM system w: $\mT_{{hc}}^{{w}}(t)$. \red{If no markers are detected in the main loop of the algorithm, one can still estimate the target marker $\mT_{{m}}^{{hc}}(t)$ by integration (executed within the function \textit{slam$\_$integrate}):}

\begin{equation}
\label{eq:vio_integrate}
\mT_{m}^{{hc}}(t)  = \mT_{{w}}^{{hc}}(t)\mT_{{hc}}^{{w}}(t-\Delta t)\mT_{m}^{{hc}}(t-\Delta t).
\end{equation}

In Equation \ref{eq:vio_integrate}, $\mT_{{w}}^{{hc}}(t)\mT_{{hc}}^{{w}}(t-\Delta t)$ is a relative transformation of camera motion from time t-1 to t and we assume a static object. In a similar fashion, the time delay of the system $t_d$ can be computed (executed with a function \textit{delay$\_$computation}) and corrected with SLAM algorithm by kinematics:

\begin{equation}
\label{eq:vio_compensator}
\mT_{{m}}^{{hc}}(t+t_d)  = \mT_{{w}}^{{hc}}(t+t_d)\mT_{{hc}}^{{w}}(t)\mT_{{m}}^{{hc}}(t),
\end{equation}

which is executed within a function \textit{delay$\_$integrate}. \red{Note that the time delay is present in any perception system (e.g. rectifying an image), fiducial marker systems as well as the communication delays. In Equation \ref{eq:vio_compensator}, $\mT_{{hc}}^{{w}}(t)$ and $\mT_{t}^{{hc}}(t)$ are computed using SLAM and multi-marker tracking.} On the other hand, $\mT_{{w}}^{{hc}}(t+t_d)$ can be computed using linear and angular velocity estimates of SLAM, multiplied by the delay time $t_d$. These two steps have several advantages. The algorithm is robust to the found failure modes of fiducial marker systems as it copes with missing marker detection, and time delays are incorporated by using velocity signals and computed delay time. Furthermore, maximum run-time of the algorithm can be pushed upto 200Hz, which is the rate of visual-inertial SLAM estimates. The algorithm deals also with drifts of SLAM estimates by using relative motion estimates only when the marker detection is lost. \red{Note that the proposed method is simple but can be an effective way of exploiting the commodity vision sensors with SLAM modules in order to improve the robustness of the fiducial marker systems.}

\subsection{The Proposed Pipeline for Objects of Unknown Geometry}
\label{subsec:unknownobject}

\red{Whenever we cannot assume the availability of the markers, the 6D pose of the objects can be estimated using depth sensors such as a LiDAR with the point cloud registration methods. For example, within the intended industrial application, the markers cannot be used for estimating the pose of the pipe. This is because in oil and gas refineries, the pipes are often very long while their inspection points are generally unknown a-priori.} Concretely, given the incoming streams of point clouds $\mP(t)$ and the point clouds of the object $\mQ$ from a CAD model, the goal is to find the rotation and translation between $\mP(t)$ and $\mQ$. Here, the point clouds $\mP(t)$ contain the points $\vp_i(t) \forall i$ with its coordinate lying at the weighted centroid $\vc_p$. Similarly, the point clouds $\mQ$ contain the points $\vq_i(t) \forall i$ with its coordinate system defined at the weighted centroid $\vc_q$. Defining $\vc_q$ to be aligned with the coordinate system of the LiDAR $l$, the 6D pose of an object $o$ can be obtained by matching the two point clouds: $\vp_i = \mR^{l}_{o} \vq_i + \vt^{l}_{o}$. This goal of finding $R^{l}_{o}$ and $\vt^{l}_{o}$ is often formulated as an optimization problem:

\begin{equation}
\label{eq:method1:plscan}
\mR^{l}_{o}, \vt^{l}_{o} = \underset{\mR^{l}_{o}, \vt^{l}_{o}}{\text{arg min}} \sum_i \rho(\left \| \vp_i - \mR^{l}_{o} \vq_i - \vt^{l}_{o} \right \|),
\end{equation}

where $\rho$ is again a distance based cost function, e.g. typically a mean squared error.

Commonly, the solution is obtained by first computing the rotation, and then the translation. Centering all the points: 

\begin{equation}
\label{eq:method1:center}
   \bar{\vp}_i = \vp_i - \vc_p \quad \text{and} \quad \bar{\vq}_i = \vq_i - \vc_q \quad \text{such that} \quad \mR^{l}_{o} = \underset{\mR^{l}_{o}}{\text{arg min}} \sum_i \rho(\left \| \bar{\vp}_i - \mR^{l}_{o} \bar{\vq}_i \right \|),
\end{equation}

the goal is to find the rotation matrix that aligns the centered point clouds. Defining the correlation matrix as $\mC = \sum_i \bar{\vp}_c \bar{\vq}_c^T$ and its singular value decomposition as $\mC = \mU \Sigma \mV^T$, the rotation can be estimated by the orthogonal Procrustes algorithm, while the translation can be obtained from the weighted centroids $\vc_p$ and $\vc_q$ after rotation:

\begin{equation}
\label{eq:method1:procrustes}
\mR^{l}_{o} = \mU \mV^T \quad \text{and then} \quad \vt^{l}_{o} = \vc_q - \mR^{l}_{o} \vc_p.
\end{equation}

This assumes the correspondences between each points to be known. In practice, however, the correspondences are often not known and the Iterative Closest Point (ICP) algorithm is often used \citep{park2017colored, rusinkiewicz2001efficient, besl1992method}. Intuitively, the ICP algorithm iterates the following steps: (1) finding the closest point in the transformed point cloud for each point: $\text{min}$ $\rho(\mP, \mQ)$, (2) estimating the transformation using Equation \ref{eq:method1:procrustes}, and (3) applying the found transformation to all points and iterate all the steps until a convergence criterion is reached. As ICP algorithm is subject to local minima, ICP is often initialized by employing the global registration methods such as \citet{zhou2016fast} or using higher level features at the first step of the ICP algorithm.

\begin{figure*}
  \centering
  \includegraphics[width=1.0\textwidth]{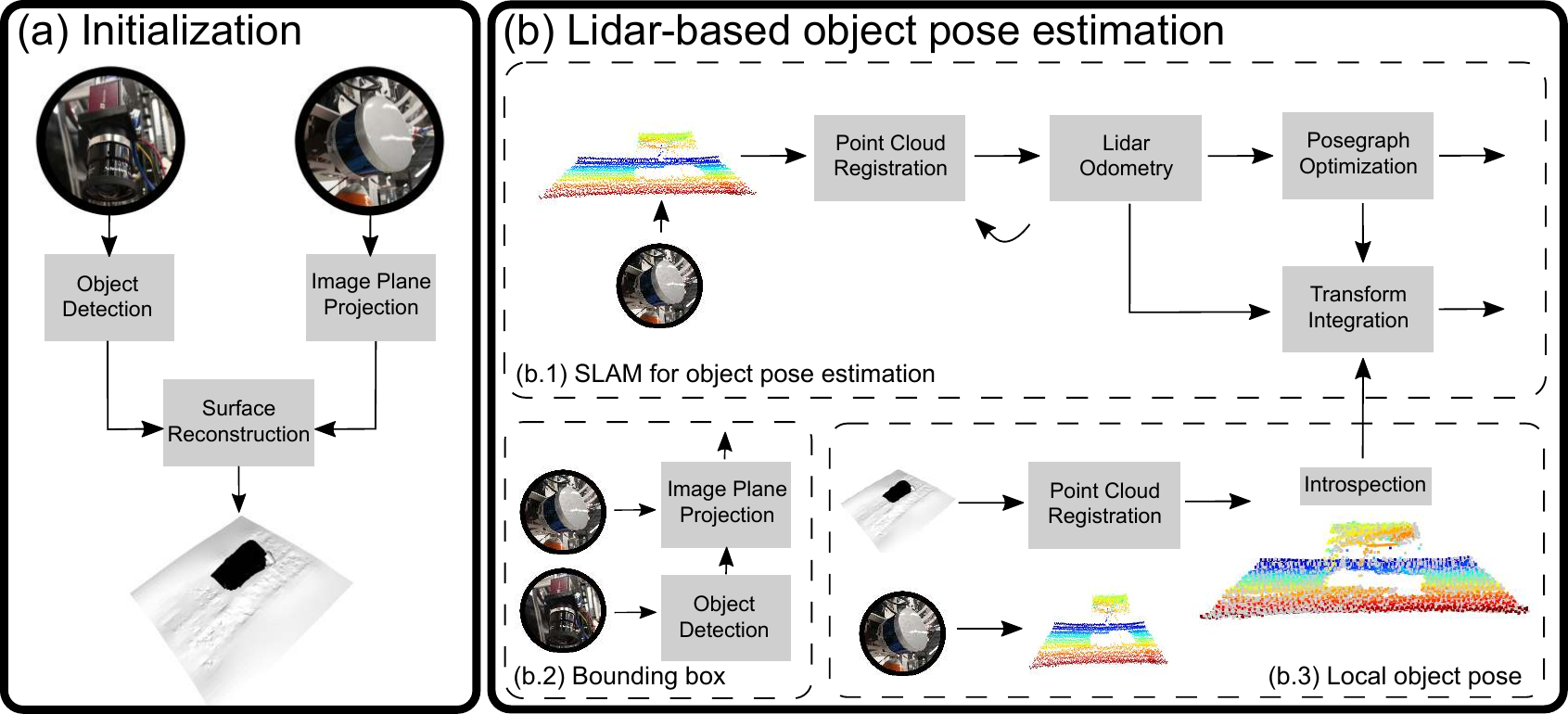}
 \caption{\red{The proposed LiDAR pose estimation pipeline.} (a) Initialization. \red{Using an object detector, the point clouds that belong to the objects of interest can be obtained for a surface reconstruction technique.} (b) Main loop. The object poses are estimated by combining a SLAM technique with local object poses from a point cloud registration method. An object detector computes the regions of occlusions and dynamic objects for masking out.}
  \label{fig:sec4:5}
\end{figure*}

\textbf{Challenges} \red{Unfortunately, such a strategy does not fully address the current use-case of the point cloud registration methods. This is because of the aerial manipulation tasks for an inspection and maintenance scenario.} We outline the resulting failure modes in Figure~\ref{fig:problem:2}. First, the strategy assumes the availability of a precisely known object geometry $\mQ$, which can be obtained through CAD models. Unfortunately, this assumption is invalid in our industrial scenario, as the CAD model of the objects that belong to external environments are unknown, e.g. CAD model of oil or gas pipe. Even though the refineries may have a 3D geometry of the site, there exist erosion and other changes to their initial model. Second, a holistic view of the object cannot be assumed. \red{In the set-up, robotic arm and other objects can occlude the object of interest, resulting in partial and overlapping view of the point cloud.} Lastly, as we deal with a floating base system where the base of a robotic arm is not fixed, view point challenges can occur. This challenges the out-of-box point cloud registration methods for LiDAR systems in Equations \ref{eq:method1:plscan}, \ref{eq:method1:center} and \ref{eq:method1:procrustes}, which contains sparse point clouds. 

\textbf{Our Solution} To this end, we propose a 6D object pose estimator using a LiDAR. The pipeline is depicted in Figure~\ref{fig:sec4:5}. For an overview, the proposed algorithm constitutes of an initialization step and a multi-process main loop. \red{At initialization, the CAD model of the scene is reconstructed online} by exploiting an object detector. \red{In the main loop,three parallel processes are created:} a bounding box estimator that computes the locations of the occluding and moving objects to be masked out, a SLAM pipeline that computes the object poses in a global reference frame, and a local object pose estimator that estimates the object poses locally. The combination of local and global methods is to mitigate the challenges related to non-holistic view of the object. The SLAM estimates can deal with large perspective changes by matching the scans sequentially, but suffers from drift. The local method, whenever is reliable, can be exploited to reduce the drift of the SLAM system. Lastly, what motivates the multi-process architecture is the efficiency, i.e., LiDAR odometry can be executed at a faster rate than the other processes that can be executed only at a slower rate.

\red{First, the pipeline is initialized by creating the CAD model of the object online. This is because one cannot always assume a known geometry of the object in the targeted application scenario, i.e., the target point cloud $\mQ$ is not available, and consequently its CAD model $\mathcal{O}$ for the VR.} Yet, from the specification of the given task, e.g. pick and place an inspection robot on a pipe, what we know a-priori is the semantics of the objects of interest, e.g. a pipe. \red{Therefore, one can still create the CAD model of the object $\mathcal{O}$ online by finding the  point clouds that belong to the objects of interest $\mP_{o}(0) \in \mP(0)$ and applying a surface reconstruction technique once}. For this, we train a DNN based object detector \citep{redmon2016you} using the eye-to-hand camera (mako). Defining this DNN as a parametric function $f_{\boldsymbol{\theta}}$ with its input as an image $\mI$, the goal of a 2D object detector is to classify and locate the objects in an image; for the object semantics $c$, e.g. $c \in \left \{ \text{pipe}, \text{robotic arm}, \text{cage} \right \}$, the classification probability $p_c$ (a score between 0 and 1), and the location as a bounding box $\vb_c \in \mathbb{R}^4$ in the given image, the 2D object detector returns the tuples: 

\begin{equation}
\label{eq:method1:dnn}
\left \{ c, p_c, \vb_c \right \} = f_{\boldsymbol{\theta}}(\mI) \quad \text{with} \quad \vb_c = \begin{bmatrix}
u_{c,1} & v_{c,1}  & u_{c,2} &  v_{c,2}
\end{bmatrix}^T.
\end{equation}

Here, the bounding box is described by two points in the image with the heights ($h=u_{c,1}$ and $h=u_{c,2}$) and the widths ($w=v_{c,1}$ and $w=v_{c,2}$) which are the top left and the bottom right corner of the box that contains the object $c$. Further defining the target object $c=o$ and using the extrinsic calibration parameter between the LiDAR and the eye-to-hand camera $\mT_{mako}^{l}$ to transform all the point clouds $\mP(0)$ to the image plane, we can find the point clouds that belong to $o$:

\begin{equation}
\label{eq:method1:pclreccrop}
\mP_o(0) = \left [ \vp_j = (x_j, y_j, z_j) \right ] \quad \text{such that} \quad j = \left \{ i \ | \ u_{o,1} \leq x_{i,image} \leq u_{o,2}, v_{o,1} \leq y_{i,image} \leq v_{o,2} \forall i \right \}.
\end{equation}

\red{This means that all the LiDAR scans $\vp_i(0)$ to the image plane are transformed, which results in $\vp_{i,image} = (x_{i,image}, y_{i,image})$.} Then, we obtain the indices $j$ of the point clouds that lie inside the bounding box of our target object $\vb_o$ and crop the original point cloud $\mP(0)$ to obtain the point clouds $\mP_o(0)$ that only contain the information about our target object. Applying a surface reconstruction technique \citep{kazhdan2006poisson}, the CAD model of the target object $\mathcal{O}$ can be created. In this way, we can still create the VR of the scenes with the objects of unknown geometry.

\red{Next, the main loop of our algorithm is described, where the first process is the bounding box estimator.} This process tackles the problem of occluding and moving objects $c=u$ by estimating the bounding box of other objects $u$ in the LiDAR coordinate system, which is for actively removing the points that belong to the occluding and moving objects $u$. \red{Concretely, similar to before, the object detector and the extrinsic calibration can be used to obtain:}

\begin{equation}
\label{eq:method1:croptwoobjects}
\begin{aligned}
\mP_{u}(t) &= \left [ \vp_u = (x_u, y_u, z_u) \right ] \quad \text{s.t} \quad u = \left \{ i \ | \ u_{u,1} \leq x_{i,image} \leq u_{u,2}, v_{u,1} \leq y_{i,image} \leq v_{u,2} \forall i \right \},
\end{aligned}
\end{equation}

where $\mP_{u}(t)$ is the point cloud that belongs to the objects $u$ at time $t$. \red{Then, the bounding boxes of the objects can be computed in xy-plane of the LiDAR coordinate system.} Defining this as $\vb^l_u$, and examining all the point clouds: 

\begin{equation}
\label{eq:method1:3dboxxz}
\begin{aligned}
\vb^l_u &= \begin{bmatrix}
\text{min}(x_u) & \text{min}(y_u)  & \text{max}(x_u) & \text{max}(y_u) 
\end{bmatrix}^T.
\end{aligned}
\end{equation}

\red{Note that the projection of all the point clouds to the image plane can be inefficient for embedded CPU computations. Therefore, a separate process is assigned. Also, moving averages and a margin are also implemented. Lastly, these bounding boxes are used to mask out the occluding and moving objects in all other modules.}

Then, in the main loop, a LiDAR based SLAM system (inspired by LOAM \citet{zhang2014loam}) is employed to address the problem of view point changes. Again, a naive strategy is to perform point cloud registration between the reconstructed CAD model of the target object $\mathcal{O}$ and the incoming point cloud scans $\mP$. However, if the initially constructed object $\mathcal{O}$ is significantly different from the current point cloud $\mP$, the point cloud registration method may perform poorly due to less overlaps between the two scans. Therefore, our key idea is that a LiDAR odometry pipeline that performs the registration between the consecutive point clouds and sets the coordinate of the constructed object $\mathcal{O}$ as a global reference, do not suffer from significant view point changes. As this approach, however, suffers from drift, i.e., accumulation of errors, two mechanisms are introduced. The first is a posegraph optimization:

\begin{equation}
\label{eq:method1:multiway}
\left \{ \mT_i \right \} = \text{arg min } \lambda \sum_i \sum_{(\mP,\mQ)\in \mathcal{K}_i} \left \| \mT_i \mP - \mT_{i+1} \mQ \right \|  ^2 + \sum_{i < j} \sum_{(\mP,\mQ) \in \mathcal{K}_{ij}} \rho(\left \| \mT_i \mP - \mT_j \mQ \right \|),
\end{equation}

where $\lambda$ determines the weight of a cost between two consecutive scans within the keyframes, and $\rho$ is a robust function, e.g. set to L2 norm in our case. \red{Here, the framework of \citet{choi2015robust} that performs robust pose graph optimization, is applied, which is less prone to the errors of pairwise registration.} Second, we propose to combine a local object poses that are obtained by performing point cloud registration of incoming scans with the target object $\mathcal{O}$. Whenever the confidence estimates of the local object poses are high (or above a specified threshold), we reset the SLAM system with initialization from the local object pose estimator. In this way, we account for the drift of the SLAM system.

\subsection{The Proposed Active Learning Framework}
\label{sec:method2}

So far, the proposed object pose estimators are described for realizing the VR based \green{telepresence} system. Here, our pipeline relied on a DNN based object detector. \red{This has been used for the online creation of a CAD model, and to rule out any occlusions and moving objects. As our entire system relies on a DNN for the VR creation, we next propose an active learning framework to obtain the required performance in DNNs within the context of field robotics.}

\textbf{Challenges} \red{The problem is on the training and the deployment of DNN based systems for various environments including both the indoor and the outdoor conditions.} The challenge lies in realizing a DNN based system for long-term operations in outdoor environments. This is due to the necessity and the manual preparations of large amounts of high quality, annotated data which can cover the variety of the operation conditions. For example, as illustrated in Figure~\ref{fig:problem:2}, a DNN trained from the annotated images of the laboratory environments, may not generalize to outdoor environments. Similarly, the seasonal variations of the scene from summer to winter can cause similar effects in deterioration of the generalization performance. Therefore, each change in the scene may require an iterative process of collecting and annotating the data. As this can be a long and tedious process, we attempt to find a principled solution that guides the process of gathering the required data for the field deployments of the DNN based systems.

\begin{figure*}
  \centering
  \includegraphics[width=1.0\textwidth]{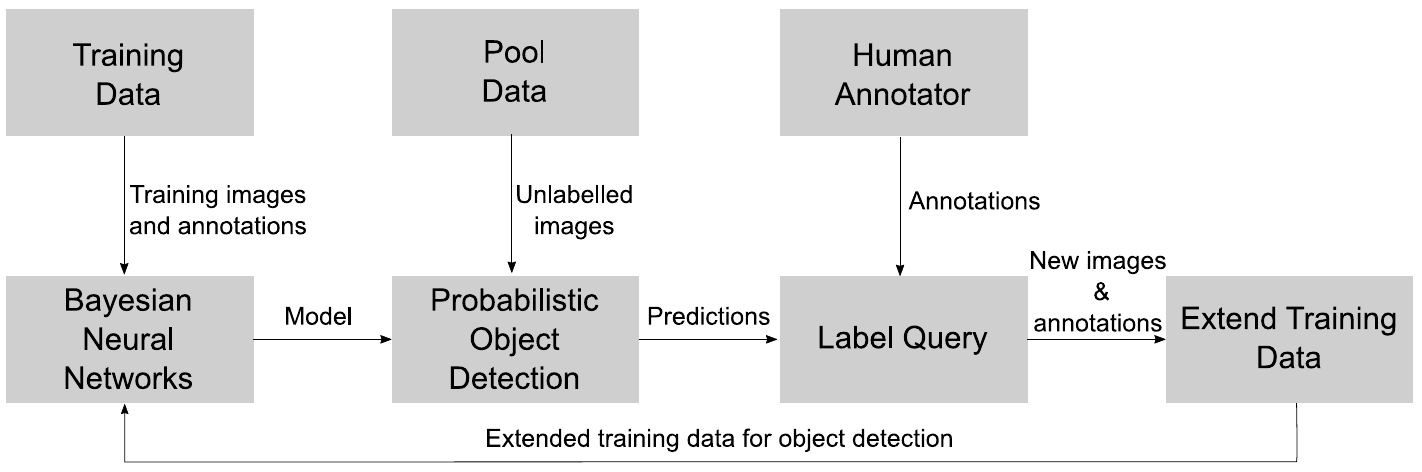}
 \caption{\red{Active learning pipeline for generating labelled data more efficiently. Instead of randomly selecting the images to be labelled, we query the most informative samples from Bayesian Neural Networks - an uncertainty-aware Deep Neural Networks for the state-of-the-art object detection frameworks.}}
  \label{fig:method2:1}
\end{figure*}

\begin{figure*}[ht!]
	\begin{center}
		\includegraphics[angle=90,origin=c,width=1.0\textwidth]{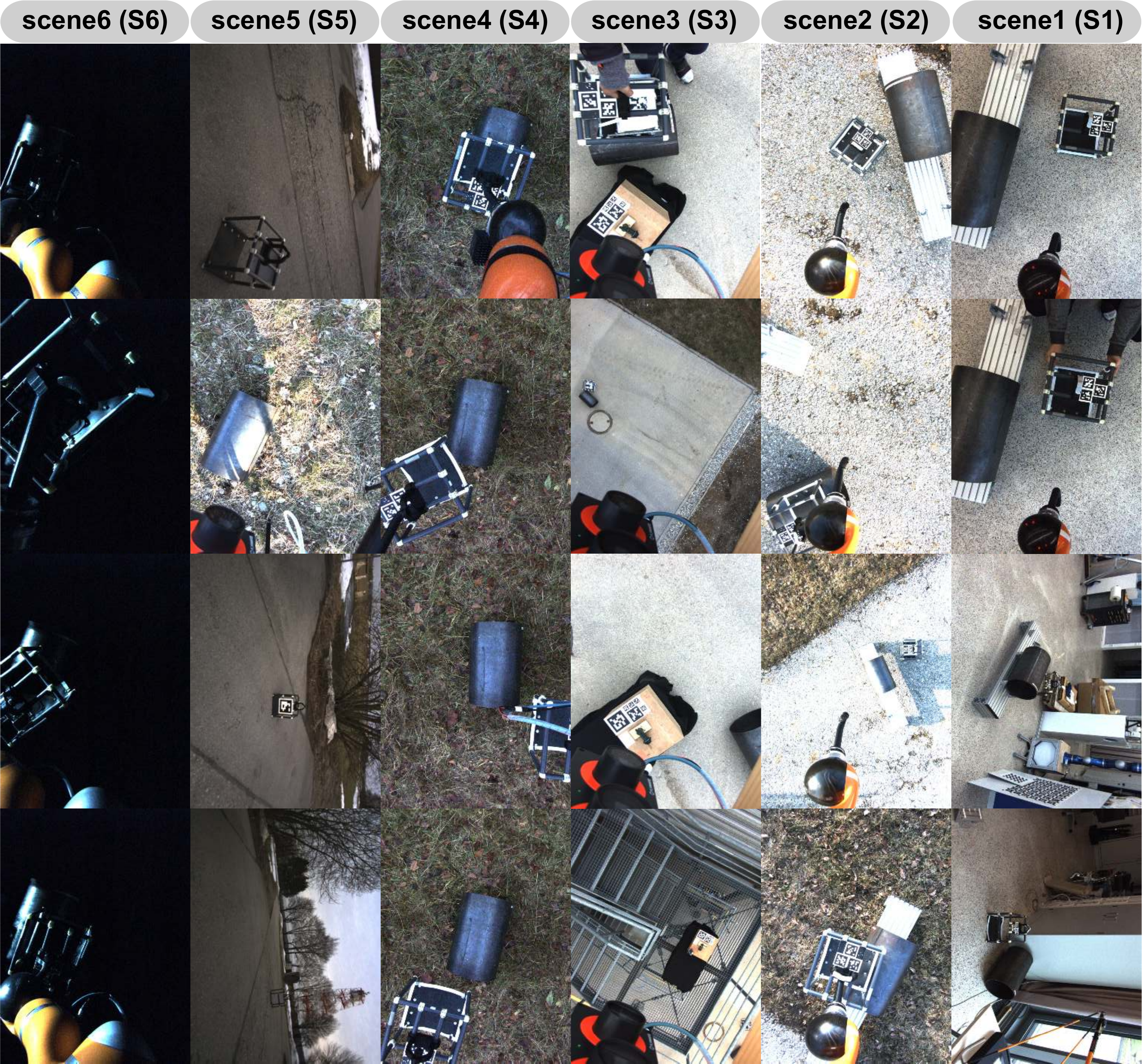}
		\caption{\red{Collected sets of pool data from different scenes. Six different scenes cover indoor environments, varying backgrounds and height, and the scenes with snow and at night.}}
		\label{fig:all::scenes:al}
	\end{center}
\end{figure*}

\begin{algorithm}
   \caption{Deep Active Learning using Bayesian Neural Networks}
   \label{alg:method1:3}
   \SetAlgoLined
   \SetKwInOut{Input}{input}
   \SetKwInOut{Output}{output}
   
   \Input{
        \begin{tabular}{l l}
        $\mathcal{D}_{init}$ & The initial annotated training data. \\
        $\mathcal{D}_{pool}$ & The unlabelled data. \\
        $Q$ & The number of query steps. \\
        $K$ & The size of the query per step. \\
        \end{tabular}
   }
   \Output{
        \begin{tabular}{l l}
        $f_{\boldsymbol{\theta}}$ & The trained DNN based object detector. \\
        $\mathcal{D}_{train}$ & The annotated training data. \\
        \end{tabular}
   }
   \BlankLine
   \Begin{
   \tcc{Initialization}
   $p(\boldsymbol{\theta}|\mathcal{D}_{init})$ $\leftarrow$ create$\_$BNN($\mathcal{D}_{init}$) \tcp*{Apply \citet{lee2020estimating, dlr137021}}
   $\mathcal{D}_{train}$ $\leftarrow$ update$\_$data($\mathcal{D}_{init}$) \tcp*{Initialize the training set from $\mathcal{D}_{init}$}
   \BlankLine
   \tcc{Main Loop}
   \For(){\emph{all the number of query steps Q}}
   {
       $p(\vy_i^\ast \mid \vx_i^\ast, \mathcal{D}_{train})$ $\leftarrow$ prob$\_$detector($\mathcal{D}_{pool}$)  $\forall i$ \tcp*{Evaluate uncertainty on a pool set (Equations \ref{eq:method1:bayesod}, \ref{eq:bma})}
       $\mathcal{D}_{selected}$ $\leftarrow$ query($\mathcal{D}_{pool}$, K) \tcp*{Query from the pool set (Equation \ref{int_acq})}
       $\mathcal{D}_{new}$ $\leftarrow$ generate$\_$annotations($\mathcal{D}_{selected}$) \tcp*{The user or human supervisor annotates the images}
       $\mathcal{D}_{train}$ $\leftarrow$ update$\_$data($\mathcal{D}_{new}$) \tcp*{Update the training set by adding new annotated data}
       $p(\theta|\mathcal{D}_{init})$ $\leftarrow$ create$\_$BNN($\mathcal{D}_{init}$) \tcp*{Apply \citet{lee2020estimating, dlr137021}}
   }
   }
\end{algorithm}

\textbf{Our Solution} To this end, we now describe the proposed pool based active learning framework \citep{cohn1996active} (also known as "experiment design"). To explain, active learning is a class of machine learning paradigm, where labelled data is not available for a supervised learning problem. Instead of obtaining annotations for all available unlabelled data, active learning attempts to only label fewer but the most informative data. Intuitively, the aim of active learning is to create a learning system that chooses by itself what data it would like the user to label. As opposed to the heuristic choice of the user, active learning enables a DNN to select small amounts of data, guiding the user in the data creation process. \red{In practical applications of robotics, this indicates that first, a pool of unlabelled data needs to be collected by the robot. As we use visual learning methods, a camera set-up could replace the deployment of the robot for data collection. We note that upto this stage, the procedure is similar to a standard supervised learning settings in robotics. Then, instead of annotating all the available data manually, fewer images are then autonomously selected by the active learning algorithm. Deep learning models are trained from these fewer images, and finally deployed to the robot.}

Figure~\ref{fig:method2:1} illustrates the overall idea behind active learning. In a pool based approach, a model is trained on an initial training set, which is often small. Then, the model selects a subset of data points from a pool of unlabeled data, and asks a human to label the selected data points. The selection involves a decision making process, which is performed through the choice of an acquisition function. Based on the updated training set, a new model is trained. Repeating this process, we can reduce the amount of labeling required to train a learning system. We present such a system for DNNs, which relies on an uncertainty quantification technique for DNNs. These are namely Bayesian Neural Networks (BNNs) and probabilistic object detection. To explain, our algorithm \ref{alg:method1:3} depicts the working principle overall. Using an initial training set $\mathcal{D}_{init}$, we train a BNN which is denoted as $p(\boldsymbol{\theta}|\mathcal{D}_{init})$. Then, for a user specified number of query steps $Q$, we first select the most informative, top K samples from the pool of unlabelled images: $\mathcal{D}_{pool}$. This is achieved by estimating the uncertainty from BNNs (denoted $p(\vy_i^\ast \mid \vx_i^\ast, \mathcal{D}_{train})$). Then, we label the selected images, and the BNN is updated with the new training set. For more explanation in detail, we next present each of these components namely the BNNs, the uncertainty of BNN based object detector, and query step through the acquisition function.

\subsubsection{Bayesian Neural Networks for Uncertainty Quantification}

One of the crucial components of the proposed active learning framework is BNNs. Intuitively, BNNs are Bayesian reasoning applied to DNNs which allows for the uncertainty quantification in DNN predictions. We note that our previous works on BNNs \citep{lee2020estimating, dlr137021} are being extended to active learning framework for object detection. While we refer to \citet{lee2020estimating, dlr137021} for in-depth treatment, our description below aims to provide the basic formulation within the context of its application to active learning.

\red{For this, consider a supervised learning} set-up with input-output pairs $\data = \begin{Bmatrix} \inputs, \outputs \end{Bmatrix} = \begin{Bmatrix} (\vx_i, \vy_i ) \end{Bmatrix}_{i=1}^{N} $, where $\vx_i \in \mathbb{R}^D$, $\vy_i \in \mathbb{R}^K$. Similar to previous sections, we define a DNN as a parametrized function $f_{\boldsymbol{\theta}}: \mathbb{R}^D \to \mathbb{R}^K$, where $\boldsymbol{\theta} \in \mathbb{R}^P$ is a vectorized form of all DNN weights or parameters, e.g. all the weights of convolution kernel or the weights and biases of a multi-layer perceptron. In a standard DNN, we typically aim to minimize the loss function: $\min_{\boldsymbol{\theta}} \quad \frac{1}{|\data|} \sum_{(\vx,\vy) \in \data} \Ls(f_{\boldsymbol{\theta}}(\vx), \vy) + \frac{\delta}{2} \boldsymbol{\theta}^T\boldsymbol{\theta}$ where $\delta$ is an $L_2$-regularizer, and $\batch \subset \data$ denote mini-batches. The resulting solution is a single hypothesis of a local maximum-a-posteriori (MAP) solution $\thetamap$. To the contrary, BNNs explicitly express DNNs as probability distributions over DNN model parameters $\boldsymbol{\theta}$ given the data $p(\boldsymbol{\theta}|\vx,\vy)$, which is also known as the posterior distribution over the DNN model parameters:
\begin{equation}
p(\boldsymbol{\theta}|\vx,\vy) = \frac{p(\vy|\vx,\boldsymbol{\theta})p(\boldsymbol{\theta})}{p(\vy|\boldsymbol{\theta})} = \frac{p(\vy|\vx,\boldsymbol{\theta})p(\boldsymbol{\theta})}{ \int p(\vy|\vx,\boldsymbol{\theta})p(\boldsymbol{\theta})d\boldsymbol{\theta}}.
\end{equation}

\red{As a direct application of Bayes theorem, where a prior distribution over the model parameters $p(\boldsymbol{\theta})$ is specified, along with the likelihood $p(\vy|\vx,\boldsymbol{\theta})$ and the model evidence $p(\vy|\boldsymbol{\theta})$.} Once the posterior distribution over the weights is obtained, the prediction of an output for a new input datum $\vx^*$ can be obtained by marginalizing the likelihood $p(\vy|\vx,\boldsymbol{\theta})$ with the posterior distribution. This step is called Bayesian Model Averaging, which can be used for active learning:
\begin{equation}
\label{eq:bma}
p(\vy^*|\vx^*, \data) = \int p(\vy^*|\vx^*,\boldsymbol{\theta}) p(\boldsymbol{\theta}|\data)d\boldsymbol{\theta}.
\end{equation}
This indicates that the uncertainty estimates for a DNN prediction $\vy^*$ can be obtained through combining different hypotheses of model parameters, resulting in the predictive distribution $p(\vy^*|\vx^*, \data)$. Another implication of the formulation is the reliance on posterior probabilities $p(\boldsymbol{\theta}|\data)$ for uncertainty quantification.

Unfortunately, estimating the weight posterior is a challenging task, and has been one of the central topic in research of BNNs \citep{gawlikowski2021survey}. While the reasons are multitude, one of the primary reasons is the lack of a closed form solution due to the nonlinearities of DNNs that prohibit the validity of conjugate priors \citep{christopher2016pattern}. As a result, the use of approximation techniques of Bayesian inference such as variational inference or Monte-Chain Monte-Carlo (MCMC) sampling have been researched with a focus on dealing with the high dimensionality of DNN weight space and the scalability with respect to large amounts of data that DNNs typically assume. \red{For the computations of the weight posterior, the proposed pipeline relies on the approaches of \citet{lee2020estimating, dlr137021}. These works are well suited for active learning in robotics, due to the demonstrated scalability to large architectures and dataset \citep{lee2020estimating}.} The extension of \citet{lee2020estimating} in the automation of the hyperparameter tuning via Bayesian Optimization \citep{dlr137021} can also be exploited in every query steps of active learning. 

\subsubsection{Uncertainty Estimation for Object Detectors}

\red{Having obtained the posterior probabilities of BNNs, the uncertainty estimates can now be computed for the underlying object detector.} A key challenge is the adaption of BNNs for the object detector that may rely on several post-processing steps \citep{harakeh2020bayesod}. As we use an anchor based detectors such as Retinanet \citep{lin2017focal} (as these types of object detectors can provide real-time performance on the Jetson TX2 as oppose to regional-proposal approaches or end-to-end pipelines), \red{one needs to deal with miss-correspondence between the anchor predictions and final outputs, and (ii) hard cut-off behavior in non-maximum suppression (NMS) step \citep{lin2017focal}. }

\red{For these, the BayesOD framework \citep{harakeh2020bayesod} is employed, which infers the output distributions from the BNNs predictive distributions.} In BayesOD, the samples of the BNNs predictive distributions are clustered in anchor level in order to derive the uncertainty estimates of the object detection. For this, one can assume that the clusters contains $M$ number of anchors. We further assume the highest classification score as the center of this cluster (indexed by 1) and other anchors of the cluster are considered as measurements to provide information for the center (denoted as $\hat{\mathbf{c}}_i$ and $\hat{\mathbf{b}}_i$). Then, the uncertainty estimates for classification $p_{[\hat{\mathbf{c}}_1, ..., \hat{\mathbf{c}}_M]}(\mathbf{c}|\vx^\ast, \mathcal{D}_{train})$ and regression $p_{[\hat{\mathbf{b}}_1, ..., \hat{\mathbf{b}}_M]}(\mathbf{b}|\vx^\ast, \mathcal{D}_{train})$ are:
\begin{equation}
\label{eq:method1:bayesod}
\begin{aligned}
p_{[\hat{\mathbf{c}}_1, ..., \hat{\mathbf{c}}_M]}(\mathbf{c}|\vx^\ast, \mathcal{D}_{train}) & \propto 
p_{\hat{\mathbf{c}}_1}(\mathbf{c}|\vx^\ast, \mathcal{D}_{train}) \prod_{i=2}^{m}p(\hat{\mathbf{c}}_i|\mathbf{c}, \vx^\ast, \mathcal{D}_{train}), \\
p_{[\hat{\mathbf{b}}_1, ..., \hat{\mathbf{b}}_M]}(\mathbf{b}|\vx^\ast, \mathcal{D}_{train}) & \propto 
p_{\hat{\mathbf{b}}_1}(\mathbf{b}|\vx^\ast, \mathcal{D}_{train}) \prod_{i=2}^{m}p(\hat{\mathbf{b}}_i|\mathbf{b}, \vx^\ast, \mathcal{D}_{train}),
\end{aligned}
\end{equation}
where $p_{\hat{\mathbf{c}}_1}(\mathbf{c}|\vx^\ast, \mathcal{D}_{train})$ indicates the per-anchor predictive distribution of the cluster center and $\prod_{i=2}^{m}p(\hat{\mathbf{b}}_i|\mathbf{b}, \vx^\ast, \mathcal{D}_{train})$ is the likelihood term. We refer to more details on the BayesOD in \citet{harakeh2020bayesod}.

\subsubsection{Acquisition Functions for Query Generation}

Another crucial component of active learning is the acquisition function, which relies on the uncertainty estimates from the BNN based object detector, in order to rank the images in the pool set. In other words, the defined acquisition function uses the uncertainty estimates to evaluate how informative each images in the pool set are. In case of an object detector, as there could be several object instance in an image, the information scores for each detected instances within an image are aggregated into one final score. Once such scores are obtained for all the images in the pool set, the top K images can be queried for the human annotation, which is then stacked into the training set. The model is then retrained with the new and larger training set, and the process repeats. As the acquisition function is a selection mechanism of active learning, its design can influence the performance of the learning framework.

\red{Within a BNN based object detector, the uncertainty estimates can be obtained for both the classification and the bounding box regression \citep{feng2021review}.} Hence, one of the design choices are on how to effectively combine the two different types of uncertainty measures - one on semantic uncertainty, and the other on spatial uncertainty \citep{feng2021review}. Defining the combination function $comb(\cdot)$ as a weighted sum or max operation \citep{choi2021active}, and also the aggregation function $agg(\cdot)$ as a sum or average operation \citep{roy2018deep}, \red{it is chosen;}:

\begin{equation}
\label{int_acq}
	\mathcal{A}(\vx_k) = agg_{j\in N_k} \left ( comb(\mathcal{U}_{j, cls}, \mathcal{U}_{j, reg}) \right ),
\end{equation}

where $\mathcal{U}_{j, cls}$ and $\mathcal{U}_{j, reg}$ are the information score of the $j$-th detection instance on an image, for the classification and the regression tasks respectively. A mechanism of this acquisition function is to first combine both the semantic and spatial uncertainty by either a weighted sum or maximum operation, and then sum or maximize over the combined score per detection instances. What motivates the given choice is the handling of the problem itself. The combination operation are to deal with having to combine the two different tasks per instance of an object detector, and the aggregation operation are to handle the multiple instances in an single image \citep{feng2021bridging}. What remains is then the design of the information scores for both classification and regression tasks: $\mathcal{U}_{j, cls}$ and $\mathcal{U}_{j, reg}$ respectively. Then,
\begin{equation}
\label{eq:method1:acquisitionfunction}
\mathcal{U}_{j, cls} = \sum_{i=1}^{|\mathcal{C}|}\mathcal{H}\left (p(c_i|\vx^{\ast}, \mathcal{D}_{train})\right ) \quad \text{and} \quad  \mathcal{U}_{j, reg} = \mathcal{H}\left (p(\mathbf{b}|\vx^{\ast}, \mathcal{D}_{train})\right ),
\end{equation}
which rely on the Shannon Entropy measure $\mathcal{H}(\cdot)$ - an indicator of how uncertain a distribution is. In case of classification, we assume categorical distributions over the classes $c_i$, while we assume multivariate Gaussian distributions for the bounding box regression $\mathbf{b}$. Importantly, what motivates for optimizing the given entropy measure is its equivalence to maximizing the information gain of a model \citep{mackay1992information}.

\section{Experiments and Evaluations}

In this work, a VR based \green{telepresence} system is proposed, which is to provide real-time 3D displays of the robots' workspace and also a haptic guidance to a human operator. The main contribution is the realization of such a system using robotic perception and active learning methods. This section therefore evaluates the proposed pipelines by examining how the created VR can match the real remote scenes, and if the identified challenges (in Figure~\ref{fig:problem:1}) are addressed by the proposed pipelines. \red{Then, the results from the field experiments are presented, in order to characterize the effectiveness of the overall system in advancing aerial manipulation for real world applications.}

\subsection{Ablation Studies and Evaluations}
\label{sec:evaluation}

\red{Several ablation studies are provided for the insights behind the presented algorithms. In particular, empirical evaluations of the devised algorithms, when facing the outlined challenges in Figure~\ref{fig:problem:1}, are the aim.}

\subsubsection{Visual Object Pose Estimation for Known Objects}

\begin{figure}
\begin{center}
    \includegraphics[width=1\textwidth]{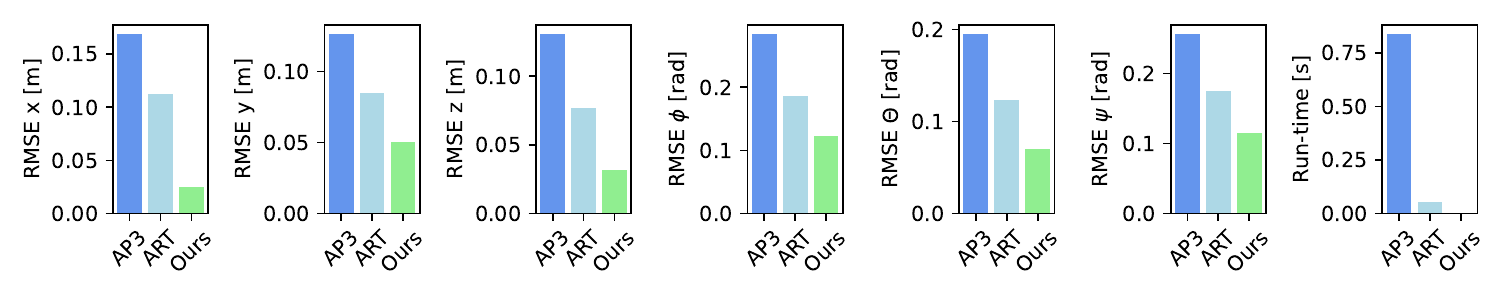}
    \caption{A summary statistics. Root Mean Squared Error (RMSE) and run-time are reported for the baseline methods and the proposed extension to the ARToolKitPlus. Lower the better for both the measures.}
    \label{rmse}
  \end{center}
\end{figure}

\begin{figure}
\begin{center}
    \includegraphics[width=1\textwidth]{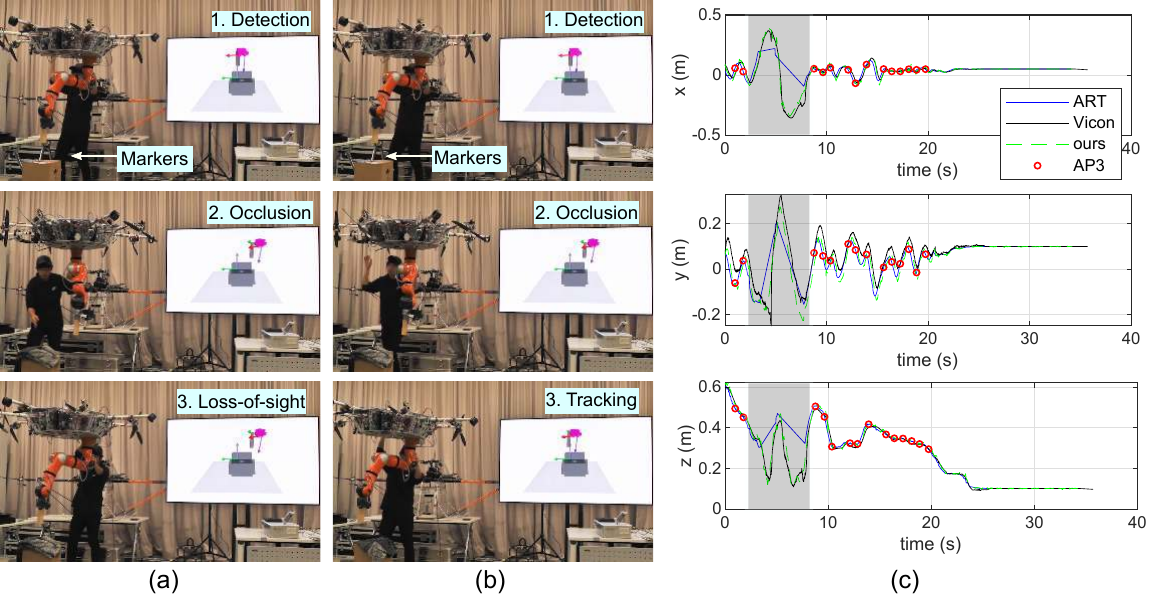}
    \caption{\red{Qualitative results of the proposed marker tracking algorithm. \red{(a) The existing marker tracking algorithms under the loss-of-sight of the markers are evaluated. (b) The proposed extension of the marker tracking algorithm with SLAM estimates is evaluated under the same scenario of the loss-of-sight. (c) The estimated positions from the baselines and the proposed algorithm are compared.} These results indicate that the proposed algorithm can cope with the loss-of-sight of the markers and the time-delays, thereby justifying the design choices of the algorithm.} \red{Three markers of size 2.5cm, a marker of 6.25cm and a marker of 10cm are used in this evaluation scenario.}}
    \label{vicon_results}
  \end{center}
\end{figure}

To recap, the marker tracking algorithms can be exploited for creating the VR with the known sets of objects. Here, the identified challenges are the shadows, the loss-of-sight or the partial views of the markers, which can cause the mismatch between the real remote scene and the VR. To address these challenges, \red{the SLAM estimates of commodity visual-inertial sensors have been integrated, and here, validation of the devised algorithm is performed.} To this end, the accuracy, the run-time and the robustness of the proposed algorithm are examined.

\textbf{Experiment Setup} \red{For this, the ground truth of the relative poses between the objects and the camera are measured using a Vicon tracking system. Then, the algorithms are evaluated on the sequences that simulates the peg-in-hole insertion tasks.} The Vicon measurements represent the ground truth of the object poses for the indoor environments. To evaluate the effectiveness of the proposed algorithm against the identified challenges, \red{the observed failure modes of the existing marker tracking systems are manually created}.  The baselines are the Apriltag3 \citep{Wang2016} (denoted as AP3) and the ARToolKitPlus \citep{artoolkit} (denoted as ART), which represent a plug-in-and-play alternatives. Particularly, as \red{the proposed algorithm} extends the ARToolKitPlus with SLAM estimates, this choice of the baseline enables a direct comparison. \red{Five repetitions of these experiments are conducted in total.}

\textbf{Results} The quantitative and qualitative results are reported in Figures \ref{rmse} and \ref{vicon_results} respectively. In Figure~\ref{vicon_results}, the estimated trajectories of the relative poses are compared with the Vicon measurements. As depicted, the proposed algorithm is robust against the lost-of-sight problems of object localization with a hand-eye camera. On the other hand, the alternatives namely AP3 and ART produce jumps as no markers are detected \red{(between t=2 to t=8 as an example) when the camera loses the sight of the markers.} This is due to the design of the algorithm \red{where the SLAM estimates of the camera pose are integrated out, whenever the markers are not detected.} Furthermore, ART suffers from a time delay, while AP3 has both the time delay, and the slow run-time. Moreover, \red{the proposed algorithm can compensate the time delay, resulting in slightly more accurate estimates.} The corresponding Root Mean Squared Errors (RMSE) is reported in Figure~\ref{rmse} along with the run-time. We observe that AP3 is slow when using high-resolution images, and this results in more errors as \red{the trajectories are compared. In the approach, these trajectories are relevant for creating the VR with object localization methods.} In this experiment, \red{the proposed} method yields the least errors and fastest run-time. We attribute the former to the robustness against the loss-of-sight of the markers, while the later is due to the integration of the inertial sensors. \red{These analysis of the accuracy, the robustness and the run-time validates the proposed algorithm. Moreover, the success of the algorithm is visually demonstrated in the video attachment, in addition to Figure~\ref{vicon_results} (a-b).}

\subsubsection{LiDAR Object Pose Estimation for Objects of Unknown Geometry}

\begin{figure}
\begin{center}
    \includegraphics[width=1\textwidth]{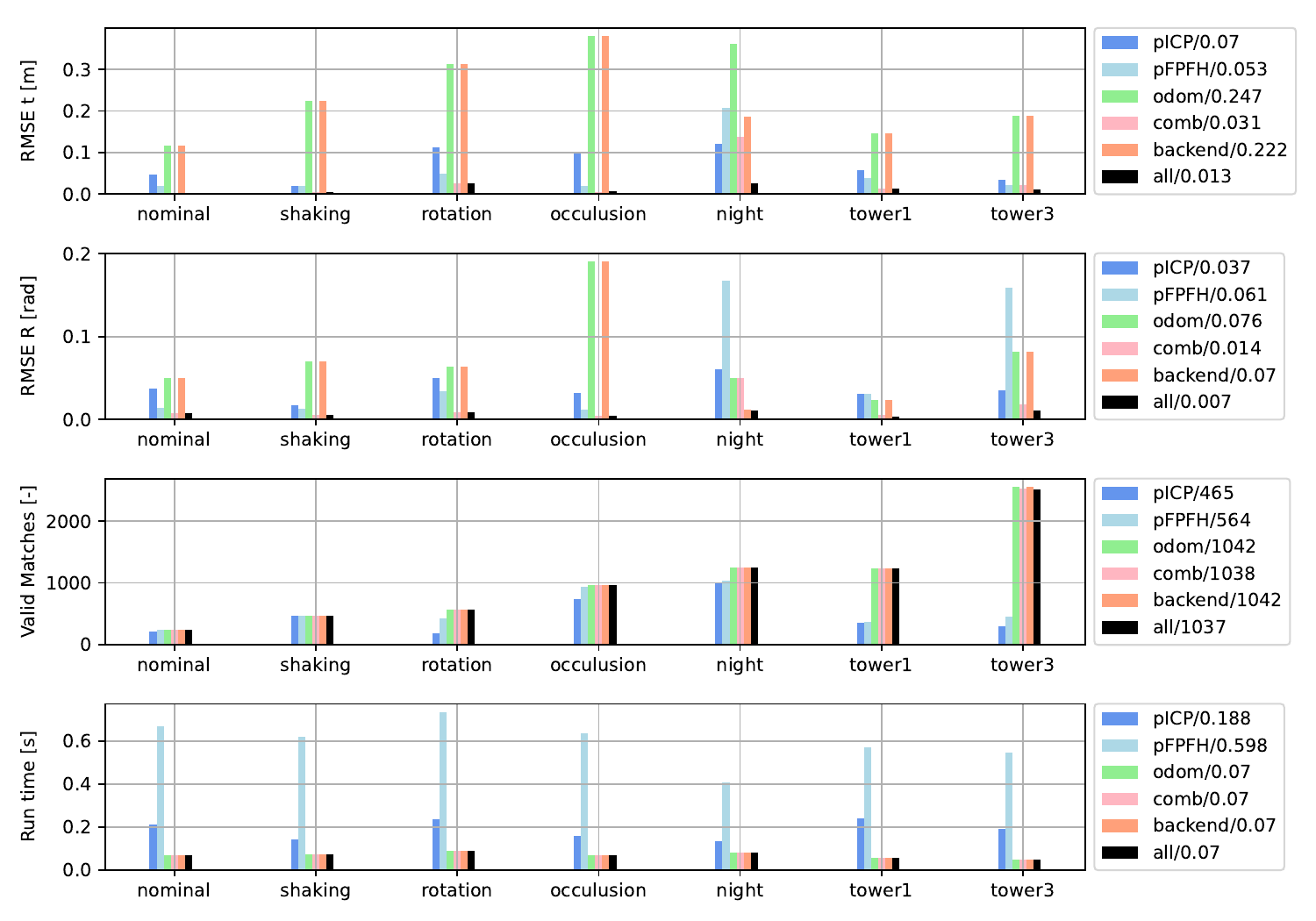}
    \caption{Statistical analysis of the baselines and the proposed method per scenario. Lower the better for the RMSE and run-time, and higher the better for the number of valid point cloud matches. These results indicate that the proposed algorithm can cope with the identified challenges, thereby justifying the design choices of the algorithm.}
    \label{vicon_results:lida}
  \end{center}
\end{figure}

For the external scenes without the availability of a precise geometry, \red{a LiDAR based object pose estimation pipeline has been proposed.} The features of the pipeline are to deal with the challenges that are outlined in Figure~\ref{fig:problem:1}. Thus, \red{the aim now is} to validate the components of the pipeline using the collected visual-inertial-LiDAR data-sets.

\textbf{Experiment Setup} For this, \red{the point clouds and the visual data are collected in various situations.} Within the controlled lab environment \red{the following scenarios are created}: a "nominal" scenario where the sensors ideally are pointed to an object statically, a "shaking" scenario in which imperfect hovering of the robot creates sensor movements, a "rotation" scenario where the robot rotates around the object, and a "occlusion" scenario in which the robot arm and other objects moves to occlude the target object. \red{These scenarios represent the identified challenges during a manipulation task (e.g. see Figure~\ref{fig:problem:1}).} To further evaluate the proposed pipeline in a realistic use-case, \red{additional scenarios are considered in outdoor environments. These are:} a "night" scenario where the sensor data were collected during a manipulation task in the night, a "tower 1" and "tower 2" scenarios where the data is similarly acquired at two different locations. Importantly, the given extensive evaluations over varying conditions are motivated by the considered industrial scenarios where \red{this paper aims to build a working system that goes beyond the proof-of-concept prototypes.}

For the baselines, \red{the off-the-shelf methods such as point-to-point ICP \citep{besl1992method, babin2019analysis}, point-to-plane ICP \citep{park2017colored, rusinkiewicz2001efficient} and the combination with the global registration methods \citep{zhou2016fast} are compared.} The pairwise registration is denoted pICP (with coarse-to-fine matching strategy) whereas pFPFH denotes the global registration method. These are to examine the vanilla object pose estimators without specific measures to address the identified challenges. For brevity, \red{only the best performing ones between the point-to-point and the point-to-plane methods are reported.} Furthermore, we compare our method without different components to evaluate the contributions of each modules to the final performance. These are the pure odometry (odom), odometry with posegraph backend (backend), combination of odometry and local object poses (comb) and the proposed object pose estimator (all). For better insights, we evaluate these methods with masked out dynamic part of the scene while existing works motivate the importance of masking out the dynamic parts of the scene in SLAM context. With these baselines, closely following \citet{babin2019analysis, park2017colored}, \red{the RMSE of the translation (RMSE t) and the rotation (RMSE R), the number of valid matches, and the run-time of each algorithms are measured.}

\begin{figure}
\begin{center}
    \includegraphics[width=1\textwidth]{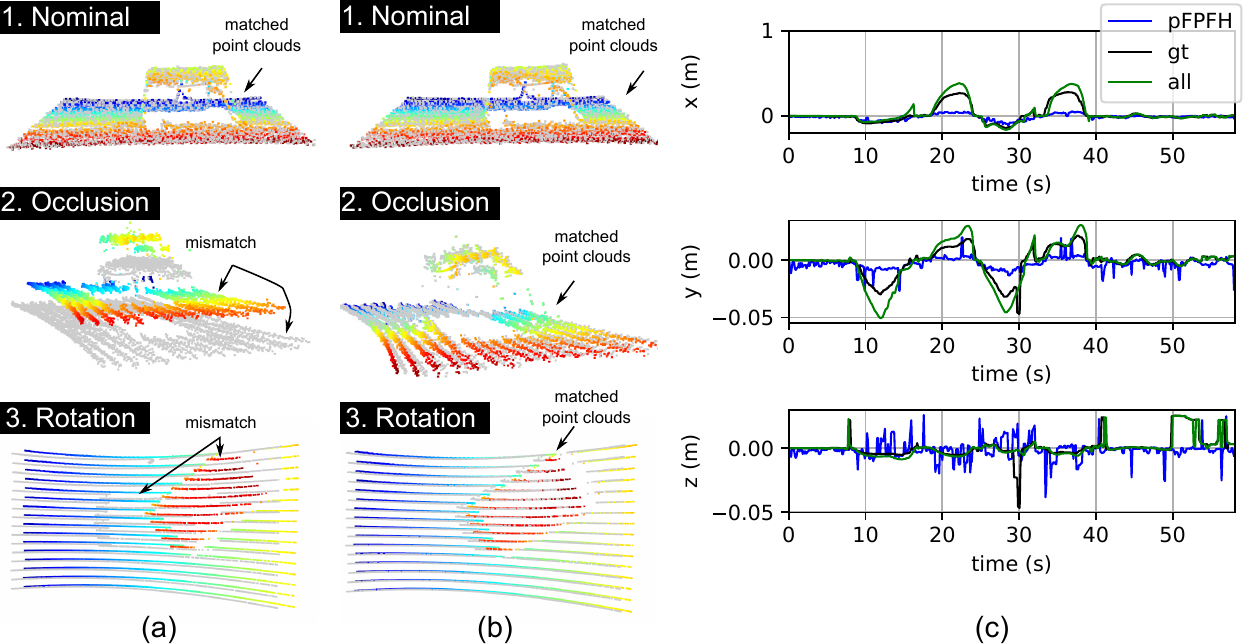}
    \caption{\red{Qualitative results for the LiDAR based object pose estimators. (a) Evaluation of a vanilla object pose estimator. (b) Evaluation of the proposed object pose estimator. (c) Plots of the estimated positions from a baseline (pFOFH) and our approach (all) against the ground truth (gt) measurements. The qualitative results over occlusion and rotation scenarios justify the design choices of \red{the proposed} method. The monotonic gray indicates the source point cloud while the colored point cloud refers to the current scan. All point clouds are cropped for the visualization purpose.}}
    \label{lidar:qualtative}
  \end{center}
\end{figure}

\textbf{Results} The quantitative and qualitative results are reported in Figures \ref{vicon_results:lida} and \ref{lidar:qualtative} respectively. In these experiments, \red{the proposed object pose estimator yielded the least RMSE for both translation and rotation.} Odometry with and without the back-end suffers from drift over time, while the vanilla methods such as pICP and pFPFH performs poorly when the view point changes are significant. The later is qualitatively shown in Figure~\ref{lidar:qualtative}, while the number of matches in Figure~\ref{vicon_results:lida} also indicate their relatively poor registration between the target and the source point clouds. \red{supports the claims of this work on the identified challenges.} Moreover, as shown in Figure~\ref{vicon_results:lida}, \red{it can be seen that all the components introduced namely pose-graph, local object poses and odometry, contributes to the accuracy of the estimates.} With respect to the run-time, the odometry based methods are real-time capable, which we attribute to the significant overlap between two consecutive LiDAR scans that helps ICP algorithm to converge faster. On the other hand, pFPFH is the slowest in terms of run-time because it relies on several components such as feature extraction, correspondence matching, and refinement through ICP. All these results are consistently observed across seven scenarios with varying degrees of severity. \red{hese experiments justify and validate the proposed methods and its design choices. Importantly, the key take-away is the effectiveness of combining object pose estimators with SLAM methods for floating-base system, which can handle the failure cases of conventional object pose estimators via introspection.}

\subsubsection{Bayesian Active Learning for Field Robotics}

\red{Lastly, the proposed Bayesian active learning framework is evaluated within the context of field robotics. To recap, the main challenge is to deal with the large variations in the environment, which may hurt the performance of an object detector that has never seen such data in the training set.} The natural question to evaluate here is the amount of labeling efforts that the active learning framework can save. As an application of the Bayesian active learning paradigm for field robotics, we focus on the impact of the system performance rather than the algorithmic advances. 

\textbf{Experiment Setup} To this end, \red{the visual data in various locations and conditions have been gathered.} These are not only the (i) laboratory environments, but also (ii) the outdoor environments in different locations. These environments are denoted as scenes 1-6 or S1-6, which are visualized in Figure~\ref{fig:method2:1}. To evaluate the system performance, \red{the manual annotations within these images have been created.} The objects are cage, pipe and robotic arm. In total of about 20k images, we randomly label 5k images. The splits are performed at the ratio of 7.5:2:0.5 respectively to a pool, test and validation set. This is to simulate the real world scenario where the training data is initially limited (e.g. the data collected in summer, and having to test in the winter). We add uniform sampling from pool data (denoted as random) and MC-dropout \citep{gal2016dropout} as the baseline. While deep ensembles \citep{lakshminarayanan2016simple} are another popular baseline, the suitability to active learning is limited due to the excessive training time. Here, the sampling strategy chosen from \citet{feng2021bridging}, and therefore, the only difference between the baseline methods are the uncertainty estimates.

Implementation details are as follows. \red{the Pytorch implementations are used, namely the Retinanet implementation from Detectron2 \citep{lin2017focal} and the official implementation of BayesOD \citep{harakeh2020bayesod} with slight modifications for better performance.} These modifications include the use of Bayesian inference only for the bounding box regression, instead of also applying to the classification head. The learning rate has also been tuned to obtain better convergence. \red{The monte-carlo samples of 30 are used for computing the uncertainty estimates, and the rank of 100 and 50 BO iterations are used.} The latter is applicable to only the Laplace Approximation, which was applied to all the layers in the Retinanet. On the other hand, only the existing dropout layers within Retinanet have been used for MC-dropout. Such implementations are motivated by the promise of each methods. MC-dropout assumes dropout layer, and have been popular in practice as one could make use of existing dropout layers, while Laplace Approximation can directly render every layer as Bayesian, given an already trained parameters. 

\begin{figure}
  \centering
  \includegraphics[width=1.0\textwidth]{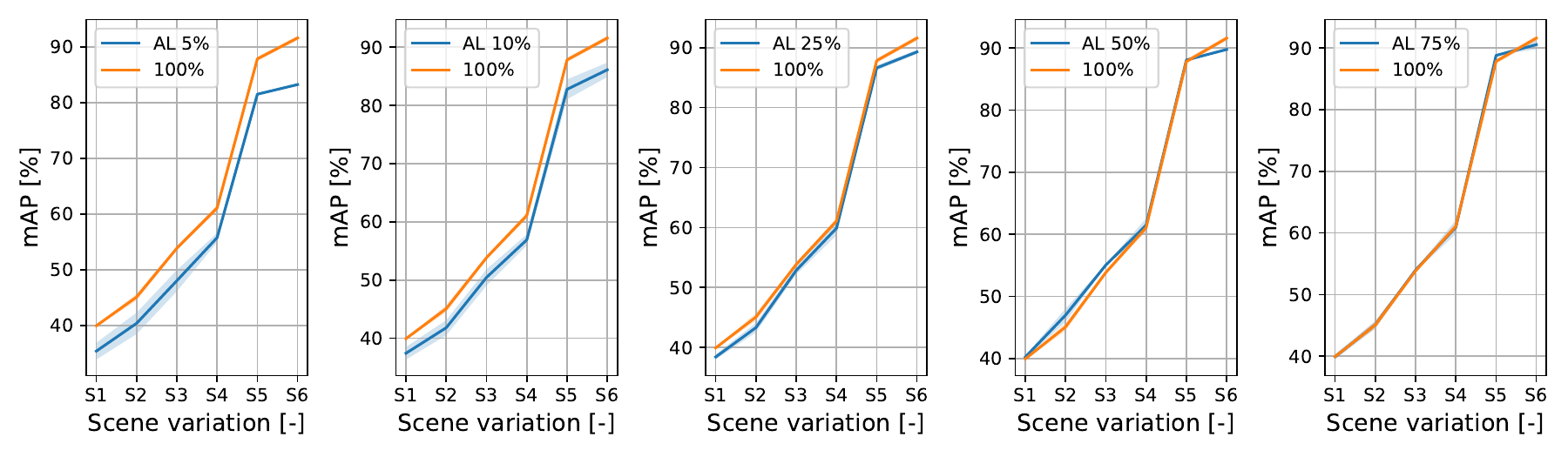}
 \caption{Results of active learning (AL) compared to the training set-up that uses 100$\%$ of all data over different scenes. AL is used for the acquisition sequence of 5$\%$, 10$\%$, 25$\%$, 50$\%$ and 75$\%$ of all data. Higher the better for mAP. \red{The results show that one can save upto 75$\%$ of data, in order to reach more than 95$\%$ of the total performance.}}
  \label{fig:al:qualtitive}
\end{figure}

\textbf{Results} \red{Firstly, it is evaluated, how much data annotations one can save by comparing the training set-ups that uses 100$\%$ of the annotated training data against the acquisition sequences of 5$\%$, 10$\%$, 25$\%$ 50$\%$ and 75$\%$ of the total data.} Repeated sequentially over each scenes, \red{the performance of the resulting object detector with mAP as a metric, are measured.} The test set contains samples from each scenes and therefore, this repetition shows how the performance gap due to scene variations are being closed. The results are depicted in Figure~\ref{fig:al:qualtitive}. We observe that the gap between the active learner (AL) and the Retinanet with 100$\%$ of annotated data (denoted 100$\%$), reduces as we increase the acquisition size from 5$\%$ to 75$\%$ of the total data. \red{In particular, in these scenarios, AL with only 25$\%$ of the total annotated data can reach more than 95$\%$ of the 100$\%$s mAP values, which results in saving upto 75$\%$ of the annotated data.} This results are due to the redundancy in the data. We believe this result can motivate AL for field robotic applications, where the data preparation can be inherently more expensive than the laboratory settings. 

\begin{figure}
\begin{center}
    \includegraphics[width=1\textwidth]{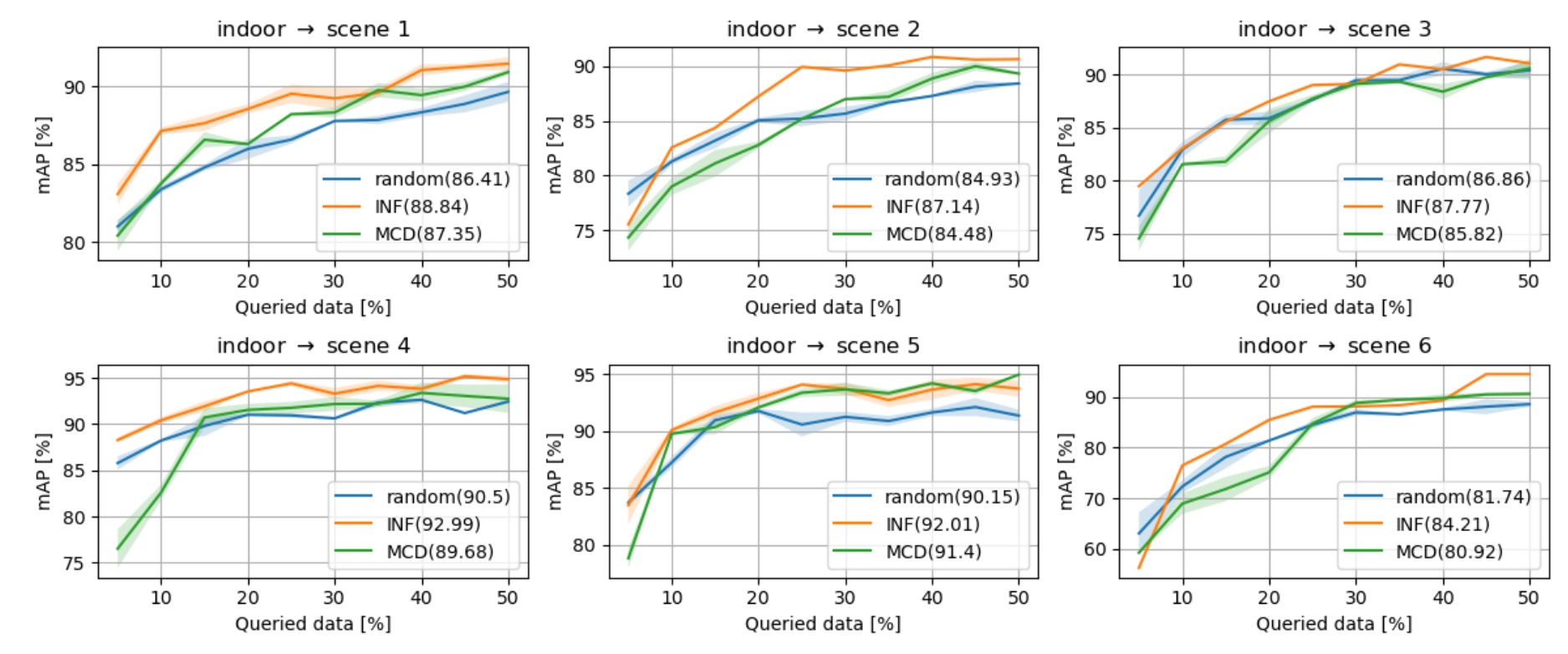}
    \caption{Comparisons of the proposed pipeline (INF) with other baselines such as random sampling (random) and MC-dropout (MCD) over six different scenes. The mean mAP during the active learning process is displayed along with the labels of the curve. The standard deviation is shown in shade. Higher the mean mAP, the better.}
    \label{fig:results:al:2}
  \end{center}
\end{figure}

Next, \red{the design choices of the proposed active learning pipeline are examined} by comparing the method against the selected baselines. The results are depicted in Figure~\ref{fig:results:al:2}. Here, \red{the transfers between the scenes are assumed}. As the robots may operate at different environments, we attempt to evaluate by starting the active learning with a neural network in an indoor scene, and acquiring the data over different outdoor scenes. For all the results, we acquire 5$\%$ of the data per iterations, and repeat 10 iterations to reach the 50$\%$ of the all data. In total, three random seeds are used to compute the standard deviation (also in Figure~\ref{fig:al:qualtitive}) for the statistical significance. Examining the mean mAP over all the iterations, the data suggests that the performance increases over using MC-dropout. The results are consistently observed in several settings with different magnitude of the improvements. We attribute the reason to post-hoc nature of our Laplace Approximation based approach. To elaborate, the methods that are based on variational inference, such as MC-dropout, might be at disadvantage in active learning settings, where each loop involves training a DNN. Naturally, as variational inference rather learns uncertainty during training, finding hyperparameters that deliver good performance over many loops is difficult. On the other hand, post-hoc methods such as ours, the uncertainty estimates are obtained after training the DNN. This decoupling enables us to extensively search for hyperparameters, which is feasible within each active learning loop. Therefore, we interpret these results to show the validity of the design choices of the proposed active learning pipeline. In summary, the key take ways are the redundancy of the data when training a neural network in dynamic and unstructured environments, and an active learning framework with well-calibrated uncertainty estimates can produce a practical and positive impact by guiding the data preparation steps towards efficiency.

\subsection{Field Testing and User Validation}
\label{sec:experiments}

\red{While the previous focus was on the validation of the methods for VR creation, the flight experiments with SAM is now presented.} The main purpose is to evaluate the benefits of the proposed system in advancing aerial manipulation capabilities. To this end, we examine two industrial scenarios in outdoor environments. \red{Then, the robustness of the proposed system is examined by varying environments and users.}

\textbf{Experiment Setup} The design of our experiment setup is to account for real world applications of aerial manipulators. \red{As a first step, the description of two industrial scenarios with SAM are provided, which involve the following aerial manipulation tasks in dynamic and unstructured environments:}

\begin{figure}
  \centering
  \includegraphics[width=1\textwidth]{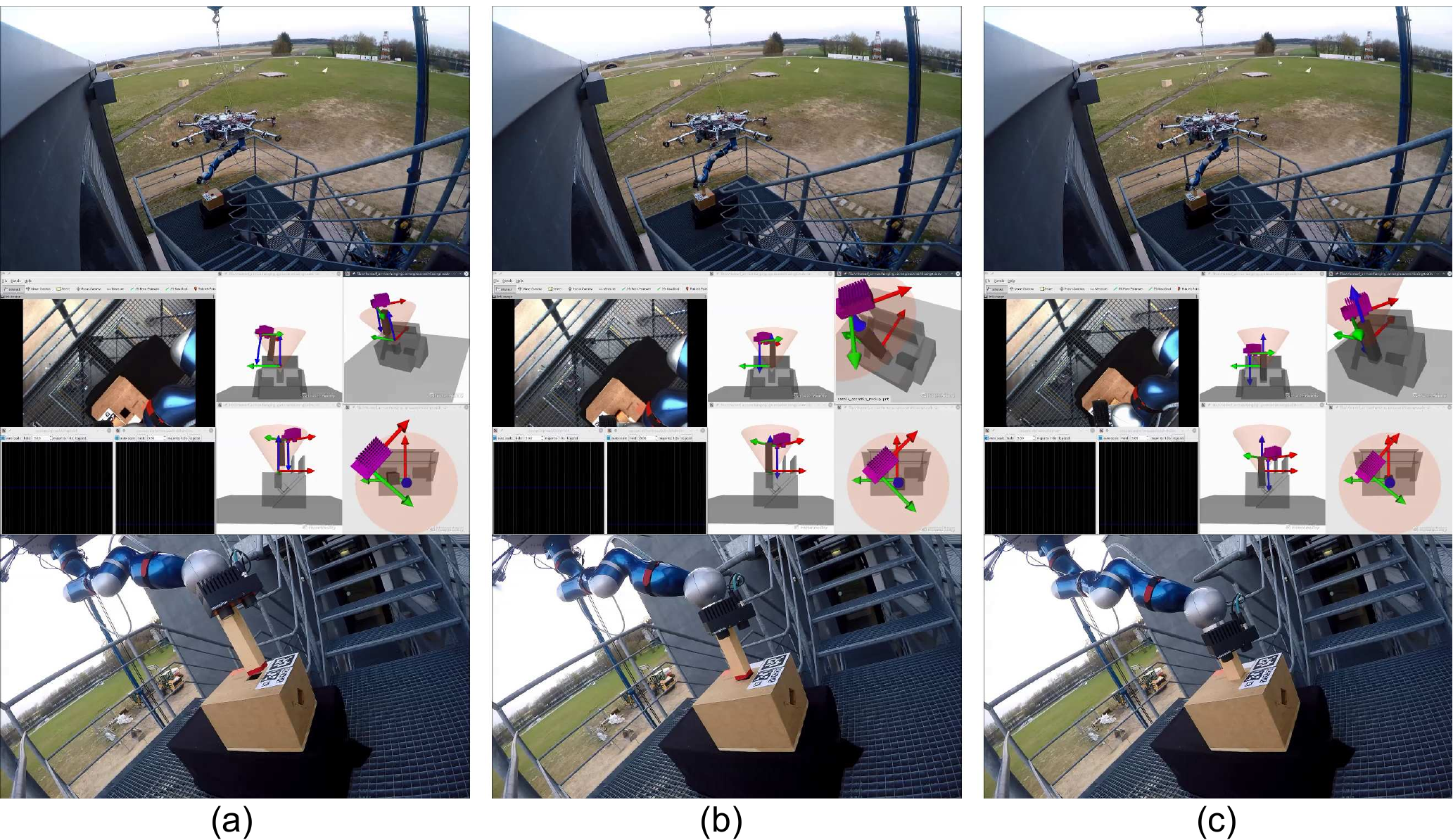}
 \caption{Qualitative results for peg-in-hole task with the aerial manipulator. (a) Approach phase, (b) precise positioning and (c) successful insertion. Top: overview of the robot's remote workspace, where SAM is suspended by a crane. Middle: the operator view with live video streams, and the created VR, which displays four different view points simultaneously, and provides haptic guidance in both position and orientation. Bottom: close view on the robot's workspace. With the proposed VR based system, the operator is able to achieve precise peg-in-hole task with the robot in outdoor environments. \red{The markers of size 2.5cm (x3), 6.25cm (x1) and 10cm (x1) are used in this scenario.}}
  \label{exp:peginhole}
\end{figure}

\begin{figure}
  \centering
  \includegraphics[width=1\textwidth]{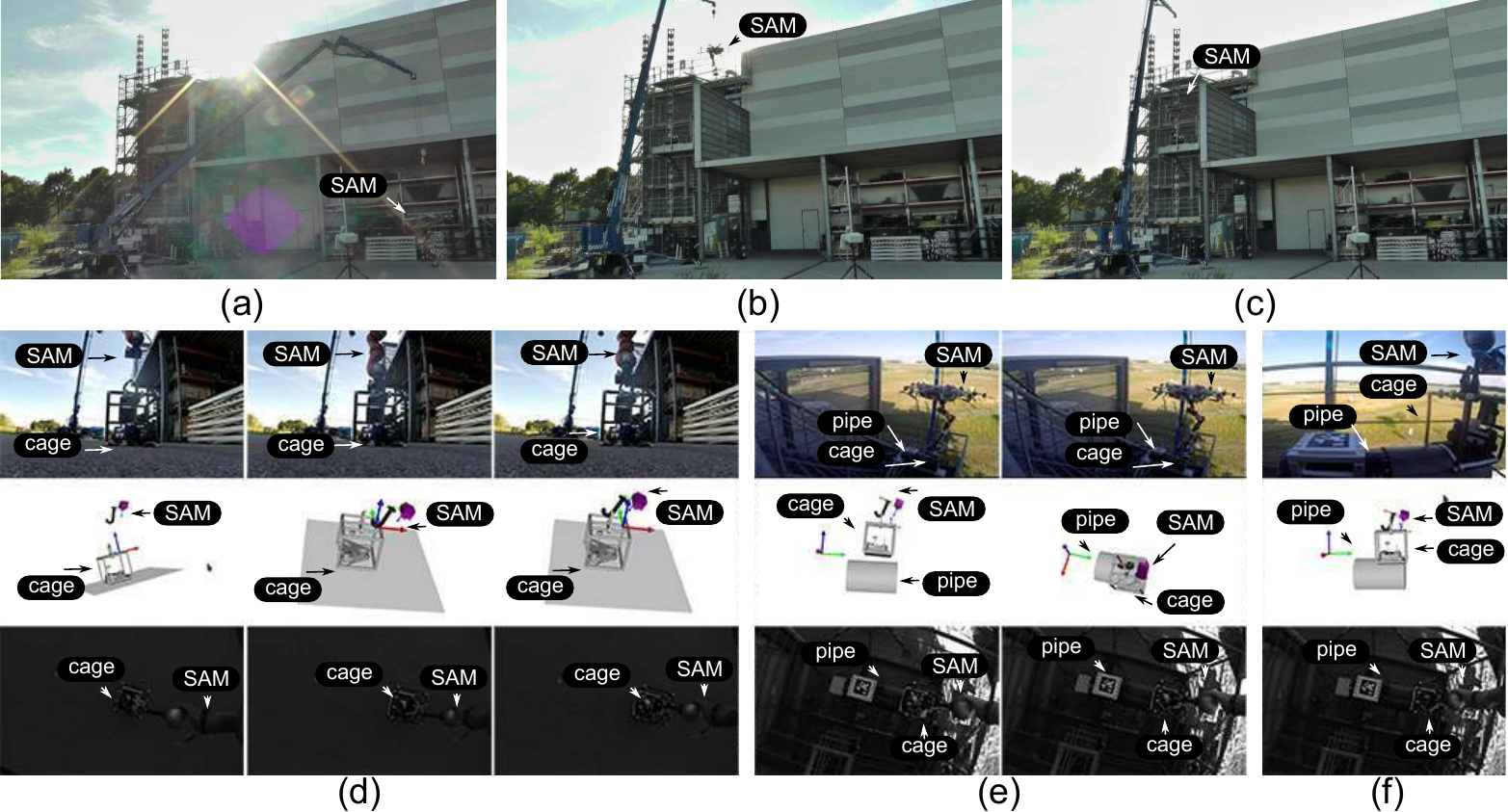}
 \caption{\green{Qualitative results for deployment of robotic crawler within an industrial inspection and maintenance scenario. (a-c) The carrier brings the robot from the ground to the remote location. (d) The robot picks the cage that carries the inspection robotic crawler. (e) The robot places the cage on a mock up of an industrial pipe. (f) The robot exerts force on the cage, so that the crawler can roll out of the cage without falling. The operator can use the VR (bottom), which contains 3D information as opposed to 2D camera images (middle). Live video stream is also subject to over and under exposure when under a shadow on a bright day. With the proposed VR based system, the robot is able to execute advanced aerial manipulation tasks for the considered real world application.} \red{Four markers of size 5cm are used for the cage hosting the mobile robot. The first experiment used a marker of 25cm on the pipe with its CAD model. Later these primitives were replaced by the proposed LiDAR based pose estimation algorithm.}}
  \label{flightexperiments:setup}
\end{figure}

\begin{itemize}[noitemsep,topsep=0pt,parsep=0pt,partopsep=0pt]
    \item \textbf{Industrial task 1: Peg-in-hole Insertion} As one of the benchmarks for manipulation, this task involves inserting an object (attached to the end-effector) into a hole. An example is depicted in Figure~\ref{exp:peginhole}. Industrial tasks such as valve opening and closing in high altitude areas, or in-air assembly of structures require the execution of this task. \red{In this work, the peg-in-hole task with an error margin of less than 2.5mm is considered.} This is a challenging task for aerial manipulation, since the robotic arm is on a floating base.
    \item \textbf{Industrial task 2: Pick-and-place and Force Exertion} \red{Two other benchmarks for manipulation are combined, which are} pick-and-place and force exertion, in the second task. In particular, our scenario, designed under the scope of the EU project AEROARMS, involves deployment and retrieval of an inspection robotic crawler.  An example is depicted in Figure~\ref{flightexperiments:setup}. It requires grasping of a cage (as a carrier of the crawler robot), placing the cage on a pipe, and pressing the cage onto the pipe while the crawler moves in and out of it for pipe inspection.
\end{itemize}

Note that, for the execution of these two tasks, the operator is located far away from the robot without direct visual contact to the workspace of the robot.  More concretely, as shown in Figures \ref{exp:peginhole} and \ref{flightexperiments:setup}, the robot operates in an outdoor environment, while its operator remotely commands the robotic arm from a ground station. This simulates a real world application scenario of a teleoperated aerial robot. 

To evaluate the feasibility and benefits of the proposed \green{telepresence} system in advancing aerial manipulation capabilities of SAM, \red{four sets of experiments are considered:}

\begin{itemize}[noitemsep,topsep=0pt,parsep=0pt,partopsep=0pt]
    \item \textbf{Set 1: Repetitions of Peg-in-hole Insertion} \red{Several repetitions of the peg-in-hole insertion task are performed} (as shown in Figure~\ref{exp:peginhole}). Here, we vary the conditions by executing the manipulation tasks with three modes, namely (a) VR and haptic guidance mode (denoted VR+HG), (b) VR mode (denoted by VR+Tele), and (c) only with live camera streams (denoted CAM). Eight repetitions are performed for each mode, and the total time for a successful execution is chosen as an evaluation metric.
    \item \textbf{Set 2: Repetitions of Pick-and-place} \red{Several repetitions of the pick-and-place task are performed} (similar to the environment in Figure~\ref{exp:peginhole} without the crane and the markers on the pipe). Here, we also consider three modes, namely (a) VR and haptic guidance mode (denoted by VR+HG), (b) VR mode (denoted by VR+Tele), and (c) only with live camera streams (denoted CAM). Six repetitions are performed for each mode, with the total time for a successful execution as an evaluation metric.
    \item \textbf{Set 3: User Validation} In a laboratory setting, \red{a user validation study is conducted with three subjects.} The variations of the users are to demonstrate that the considered manipulation tasks can be performed by different users. The considered tasks are force exertion onto a pipe for three seconds (denoted by validation task 1), and also placing a cage on a pipe with and without moving base (denoted as validation tasks 2 and 3 respectively). With VR and haptic guidance mode, the total time for a successful execution is chosen as an evaluation metric.
    \item \textbf{Set 4. Operations at Night with VR} For both the industrial tasks, we perform experiments at night without sunlight. \red{With flash light from an external source, the feasibility and benefits of the proposed system are demonstrated.} At night in outdoor environments, this functionality of being able to perform manipulation tasks is important to increase the range of operation hours including emergency services for several industrial use-cases of aerial manipulators.
\end{itemize}

\red{With these sets of experiments, the aim is to examine the following aspects. For Set 1, the VR+HG mode are examined for achieving manipulation tasks with a high precision.} The proposed marker tracking algorithm is utilized here. \red{With Set 2, the benefits of VR in providing depth information to the operator with haptic guidance only for the orientation, are examined.} The LiDAR based object pose estimator is utilized here for pipe localization, while the pose of the cage is monitored with the marker tracking method. Set 3 aims for a user validation, while with Set 4, we attempt to push the limits of the proposed system. \red{Again, the use-case of the developed system is to augment the live video streams by providing 3D visual feedback and haptic guidance. The use of VR interface only, is not the intended use-case of the system.} \red{Besides, in industrial scenarios of pipe inspection and maintenance, the pipes are often very long and their inspection points are unknown a-priori. Therefore, the proposed Lidar based pose estimation method is used to localize the pipe. This use-case justifies the assumption that the object is semantically known, but no geometry is available.}

\begin{figure}
  \centering
  \includegraphics[width=1.0\textwidth]{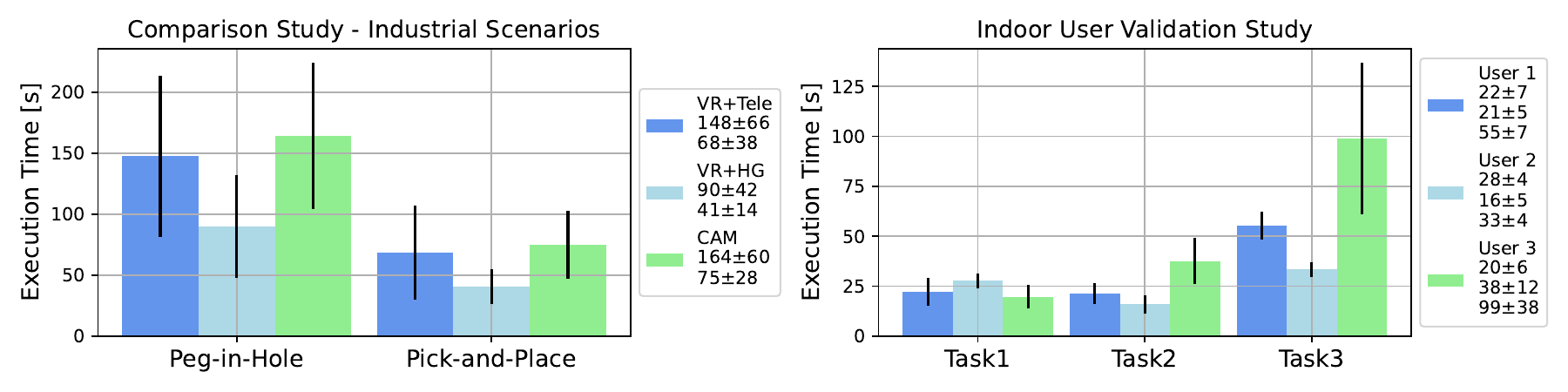}
 \caption{Performance of SAM in terms of execution time. Left: the results of the comparison study is depicted, where we compare pure VR based \green{telepresence} (VR+Tele), VR with haptic guidance (VR+HG) and a solution using only a camera (CAM). The statistics are computed over 24 and 16 successful executions in outdoor environments for the peg-in-hole and pick-and-place tasks, respectively. Right: the results of user validation is shown, where three users performed three pre-designed tasks, namely force exertion and placing the cage on the pipe. The statistics are computed over 27 successful executions in an indoor environment. Lower the execution time, better the performance.}
  \label{fig:stats:all}
\end{figure}

\textbf{Results} The results of Sets 1, 2 and 3 are depicted in Figure~\ref{fig:stats:all}. First and foremost, the comparison study of the peg-in-hole insertion tasks with 24 successful executions shows that VR+HG requires the least execution time, while VR+Tele and CAM took similar mean execution times. \red{Note that the executions with CAM used an automated initialization of the end-effector orientation, which was to make the task execution successful.} The superiority of VR+HG is expected as the human operator is assisted by the haptic guidance system. Similarly, the comparison study of the pick-and-place task shows similar trend, where the statistics are computed using 18 successful executions in an outdoor environment. As a third point, the user validation demonstrates that all the tasks can be executed by different users with different performance characteristics. Results from the scenario with a moving base, namely  validation task 3, required more time for execution, which indicates that the tasks are more challenging with a moving base. Overall, these studies indicate the performance benefits of the system including feasibility and robustness of the proposed system.

\begin{figure}
\begin{center}
    \includegraphics[width=1\textwidth]{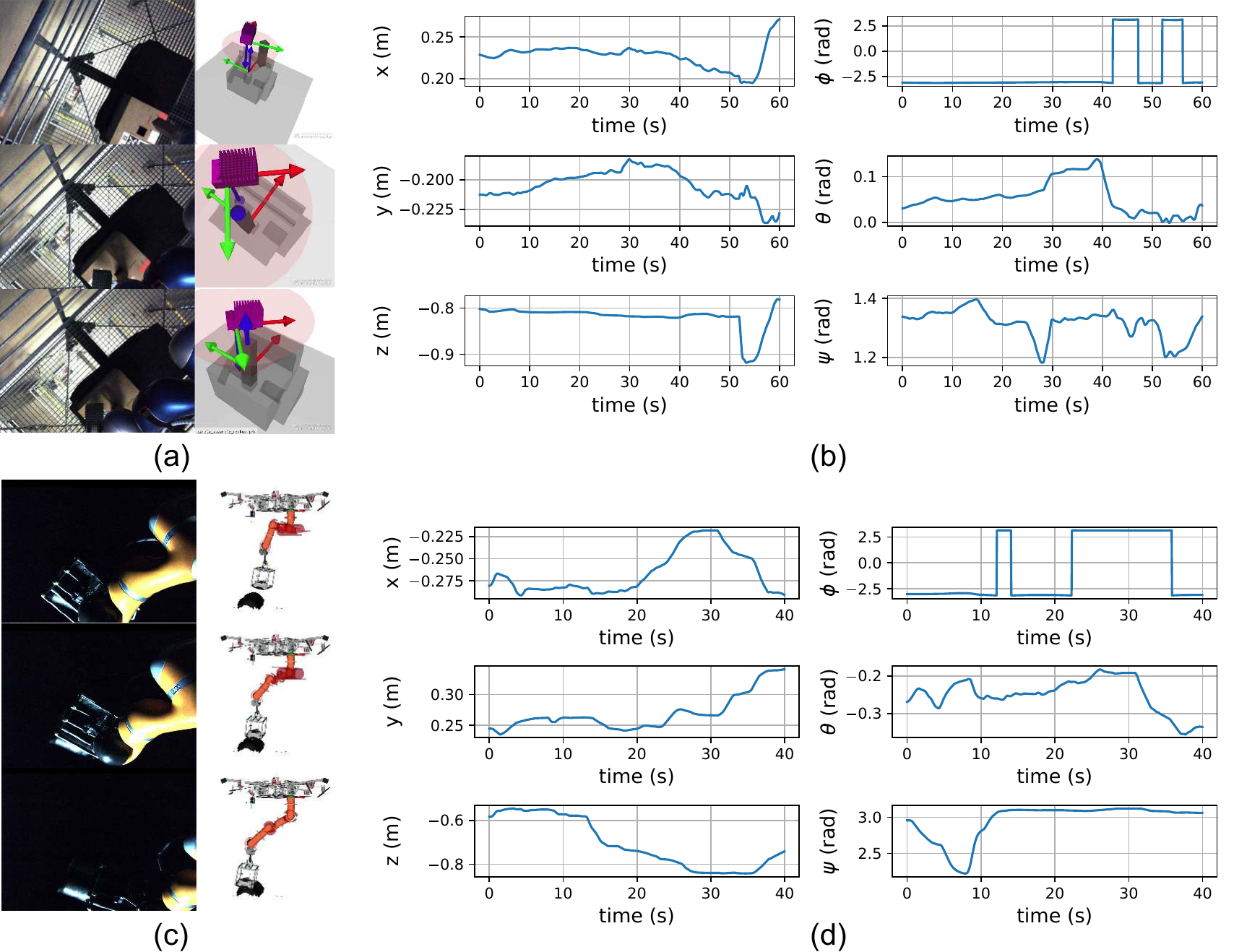}
    \caption{Qualitative visualization of aerial manipulation performed during the night. (a, c) The camera view and the VR. (b, d) The translation and the orientation of the robots' end-effector. The corresponding pairs are (a,b) and (c,d). These results suggest that the proposed system can extend the operational range of SAM, which further establishes its viability for real world industrial applications. \red{External views are depicted in Figure~\ref{fig:night:results:scene}.}}
    \label{fig:night:results}
  \end{center}
\end{figure}

The qualitative results of Set 4 are depicted in Figure~\ref{fig:night:results}. The figures show the live-stream view from the eye-in-hand camera, and also from the VR. Lights are provided from an external source and the camera exposures are tuned to achieve balance between noise, brightness and stability of streaming. Poses of the end-effector are plotted to illustrate task executions. These plots are also similar for the peg-in-hole and the pick-and-place tasks from previous sets of experiments. Notably in (b) of Figure~\ref{fig:night:results}, peg-in-hole insertion is best characterized in z-axis between 50s and 60s. \red{In (d) of Figure~\ref{fig:night:results}, the placements are observed in z-axis between 28s and 35s, while the effects of haptic guidance is shown between 8s and 15s.} These experiments show that the proposed concept can also work under the unfavorable lighting conditions, thereby extending the operation range of the aerial manipulators. \red{Additional plots for the interaction wrenches during the manipulation task executions can be found in the appendix.}

\subsection{Discussion}

The results obtained with ablation studies and field experiments suggest successful development and deployment of the proposed VR based \green{telepresence} system for advancing aerial manipulation. For providing both the sense of touch and the sense of vision to the human operator, the proposed system featured not only a haptic device, but also a VR interface that provides a real-time 3D display of the robot's workspace as well as a haptic guidance. \red{In the experiments, it is shown that the system neither requires any external sensors nor pre-generated maps, copes with the challenges of a floating-base manipulation systems,} \green{i.e., induced motions of attached sensors due to coupling between the manipulator and the base, fuses multiple sensors whenever appropriate,} and has been exhaustively evaluated outside laboratory settings. These features of the proposed VR system are requirements for several industrial applications of aerial manipulation technologies. To the best of our knowledge, using on-board robotic perception only, this work is the first of its kind to demonstrate the feasibility of such VR based concept in dynamic and unstructured environments, which includes several outdoor locations, days and nights, as well as different seasons. 

\red{To build such a system, several methodological insights are provided, from the identification of practical challenges to their working solutions, both of which are validated using the real data from robot's sensors.} In particular, the object pose estimators are subject to non-holistic view of the objects, which includes loss-of-sight, partial view and occlusions as examples. For this, we have combined the object pose estimators with ego-motion tracking of the environments using real-time SLAM estimates. In the custom data-sets that emulate these challenges, the results show that the identified problems can be coped with, which has resulted in the pipelines that meets the requirements of VR creation in accuracy, run-time and robustness. \red{Moreover, when one aims for a long-term deployment of a learning system in outdoor environments, we find that data collection and preparation become a practical problem. To this end, a pool based active learning pipeline has been evaluated, which used a previous work on uncertainty quantification \citep{lee2020estimating}.} In a field robotics settings, \red{the results show that only $25\%$ of total data is enough to reach $95\%$ of a solution with all data points and other baseline approaches can be outperformed, overall improving the sampling-inefficiency of DNNs.}

\red{Overall, the experiments of this work illustrate several benefits of the proposed VR based} \green{telepresence} system for advancing aerial manipulation capabilities in real world applications. Intuitively, a virtual environment allows the human operator to change its sight-of-view, zoom in and out, and provides a haptic guidance. \green{In the presented comparison study (with 40 task executions in outdoor environments; a single user), the results show significant reduction in the total execution time when using the proposed system with haptic guidance. The user validation study (three users with 28 total task executions) suggests that three different users can execute the tasks successfully with varying degrees of performance.} \red{Moreover, with the demonstration of the operations at night, the range of operation hours has been extended for the current aerial manipulation systems.} All these results are obtained within two industrial scenarios that requires advanced aerial manipulation capabilities, namely pick-and-place, force application and peg-in-hole, which goes beyond a contact based inspection. Therefore, these results demonstrate the viability of the proposed VR based \green{telepresence} concept for industrial applications in the real world.

\subsection{Lessons Learned}
\label{sec:lessons}

During the flight campaigns with SAM, we learned a few lessons, which we would like to share with the community. These lessons learned are centered around the proposed VR based \green{telepresence} system. \red{Note that the focus herein is on the use-cases of the proposed system, the design choices, and the limitations.}

\textbf{On Use-cases of VR with Haptic Guidance for Aerial Manipulation} The necessity of VR with haptic guidance for SAM (or robots with similar morphology) largely depends on the choice of the haptic device and difficulty of the manipulation tasks. In the initial flight experiments using only the 2-DoF Space Joystick Rjo in \citet{lee2020visual}, the operator could not easily complete the considered manipulation tasks by only relying on live camera streams. On the other hand, at the later stages of development, it was much easier for the operator to complete the tasks, when we augmented the system with the 6-DOF haptic device Lambda. With the 6-DoF device Lambda and a whole-body controller of the suspended platform to handle occlusions and enhance the camera's field of view (e.g., in \citet{9476739}), the operators could also complete the tasks using only live video streams, despite slower execution time. 

However, while the necessity of VR and haptic guidance may depend on the system and the complexity of tasks, we find that the combination of VR, haptic guidance and live video stream resulted in the best performing system. Intuitively, the live video stream can provide situational awareness to the human operator, but suffers from over- and underexposure depending on the light conditions, camera jitters due to the movement of the platform under severe winds, lack of complete 3D information and inability to provide haptic guidance. The proposed VR system can complement the live video stream as it does not degrade with outdoor conditions, provides complete 3D information with an option to change the field of view, and supports haptic guidance. Another benefit is that VR enables seeing the "full model" instead of the limited field of view of the camera at its current position, which includes configuration of the robotic arm.

\begin{figure}
	\begin{center}
		\includegraphics[width=1\textwidth]{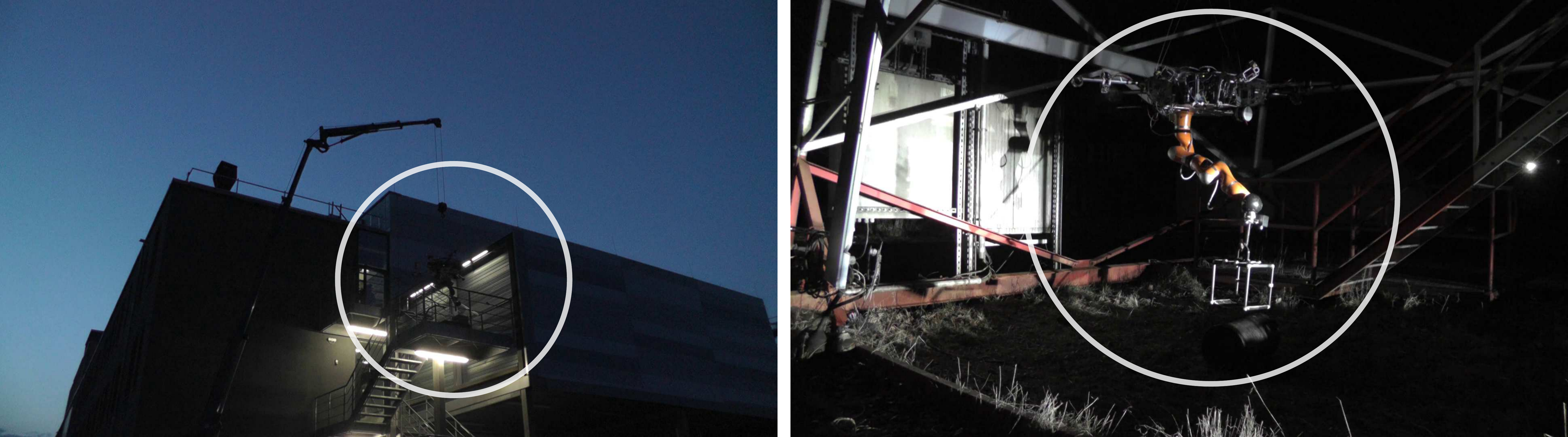}
		\caption{\red{Aerial manipulation at night. The views of the scene from an external camera are depicted. Left: SAM performing peg-in-hole insertion task at night. Right: SAM placing a cage onto a metal pipe for the deployment of a robotic crawler. SAM and the objects are highlighted in white. }}
		\label{fig:night:results:scene}
	\end{center}
\end{figure}

\textbf{On Scene Graph Verses 3D Reconstruction} The VR creation from robot perception can either rely on scene graph or 3D reconstruction techniques, where the choice of the approach largely depends on the validity of either static-base or floating-base assumptions. For example, a ground based mobile manipulators can first stop, and then perform manipulation. If the scene and objects are static, the relative motion between the sensors and the objects can be easily estimated, and the real-time capability from the perception algorithm is not required. In such a scenario, relying on the outputs of 3D sensors such as RGB-D or stereo would be the simplest option to implement. The robot can map the environments and the objects first to ensure a good field of view, e.g., avoiding occlusions, and then use the map to create a VR. On the other hand, if the relative motion between the sensors and the objects are consistently changing, e.g., in a floating-base system like ours, we find that the scene graph approach can be better suited. The scene graph approach can rely on the object pose estimators that are fast and accurate, and the existing corner cases such as occlusions and loss-of-sight can be handled by using the proposed pipelines. Another consideration is bandwidth, i.e., the object poses require only 6D vectors while streaming point clouds is more expensive. The 6D pose representation can also be plugged in directly for the shared controllers with position based visual servoing (as in VR+HG).

\textbf{On Inherent Uncertainty in VR Creation} The proposed VR from robot perception cannot match the reality perfectly. In spite of this limitation, the considered task could be successfully completed even for several challenging outdoor environments. What attributed to the successful deployment of the proposed VR system was identifying when the VR was prone to failures (see Figure~\ref{fig:problem:2}). The provided object pose estimators mitigates the identified failure cases by combining a standard object pose estimator with tracking of the environments. Here, the combination is facilitated by a module that identifies the failure cases, e.g., self-evaluation of point cloud processing methods, and missed detection of the markers while using visual-inertial systems. Moreover, in the proposed active learning pipeline, a more explicits representation of uncertainty is used to improve the data preparation steps for our DNN based component. Therefore, we find that reliability awareness of an algorithm is crucial for the robotic systems to achieve complex tasks in dynamic and unstructured environments. This is in line with \citet{thrun2000probabilistic}. 

\red{The current use of DNN's uncertainty has been off-line, like pool based active learning, while its use on-board the robot could potentially offer several more benefits.} In this regard, combining a real-time uncertainty estimation method \citep{lee2022trust} with a reliability-aware shared control architecture \citep{balachandran2020adaptive}, could be an interesting direction of future research for reliable operations of complex systems in unstructured and dynamic environments. \green{Lastly, a full-scale user study is envisioned, which is tailored on telepresence robots with aerial manipulation capabilities, in outdoor environments.}

\section{Conclusion}
\label{sec:conclusion}

In this article, \red{the real world applications of aerial manipulation in dynamic and unstructured environments are envisioned.} \red{A novel telepresence system has been proposed, which} involves not only a haptic device for the sense of touch, but also a virtual reality (VR) for enhancing the sense of vision and further providing haptic guidance. \blue{To create such system, we identified challenges while using off-the-shelf methods, and devised several extensions to address them. These techniques include pose estimation pipelines for industrial objects of both known and unknown geometries, and also a deep active learning pipeline to efficiently collect and annotate training data. Empirically, we validated the proposed methods using data-sets collected from the robot's sensors. With these, the influence of each component is examined with regard to mitigating the identified challenges, and we demonstrate the feasibility of creating the real-time and accurate VR. Methodologically, the key to success was an awareness of the algorithms' own failures and uncertainty -- also known as robotic introspection. One example is the combination of object pose estimation and SLAM, which is facilitated by a module that identified the failure cases. Another example is the active learning pipeline, where information gain is computed from an explicit representation of uncertainty. Most importantly, with the DLR's SAM platform, we conducted exhaustive experiments over extended durations in which we executed over 70 complex aerial manipulation tasks to characterize the performance of the resulting system. The obtained experimental results show that the proposed system can reduce the execution time of both pick-and-place and peg-in-hole insertion tasks by approximately 1.8 times. The system is also demonstrated to operate at night without any direct sun light. Therefore, the effectiveness of the proposed telepresence system is demonstrated for future industrial applications of aerial manipulation technology. }

\appendix

\section{Platform Design, Control, Teleoperation and IT Architectures}
\label{appendixA}

\red{In this section, we present the details about platform design and control, teleoperation system and IT architectures.}

\subsection{Platform Control}

\begin{figure}
	\floatbox[{\capbeside\thisfloatsetup{capbesideposition={right,top},capbesidewidth=4cm}}]{figure}[\FBwidth]
	{\caption{\red{The control framework of SAM. Propeller based oscillation damping and yaw controller, torque controller for robotic manipulator, and feedforward position controller for the winches. Oscillation damping and yaw controller use propeller based actuation to stabilize the platform while SAM is performing manipulation task. IMU provides feedback signals. Impedance based torque control is performed for the robotic arm, where joint encoders and torque sensing are used as feedback. The three winches can adjust the length of the cable suspending the platform.}}\label{fig:appendix:1}}
	{\centering
		\includegraphics[width=0.75\textwidth]{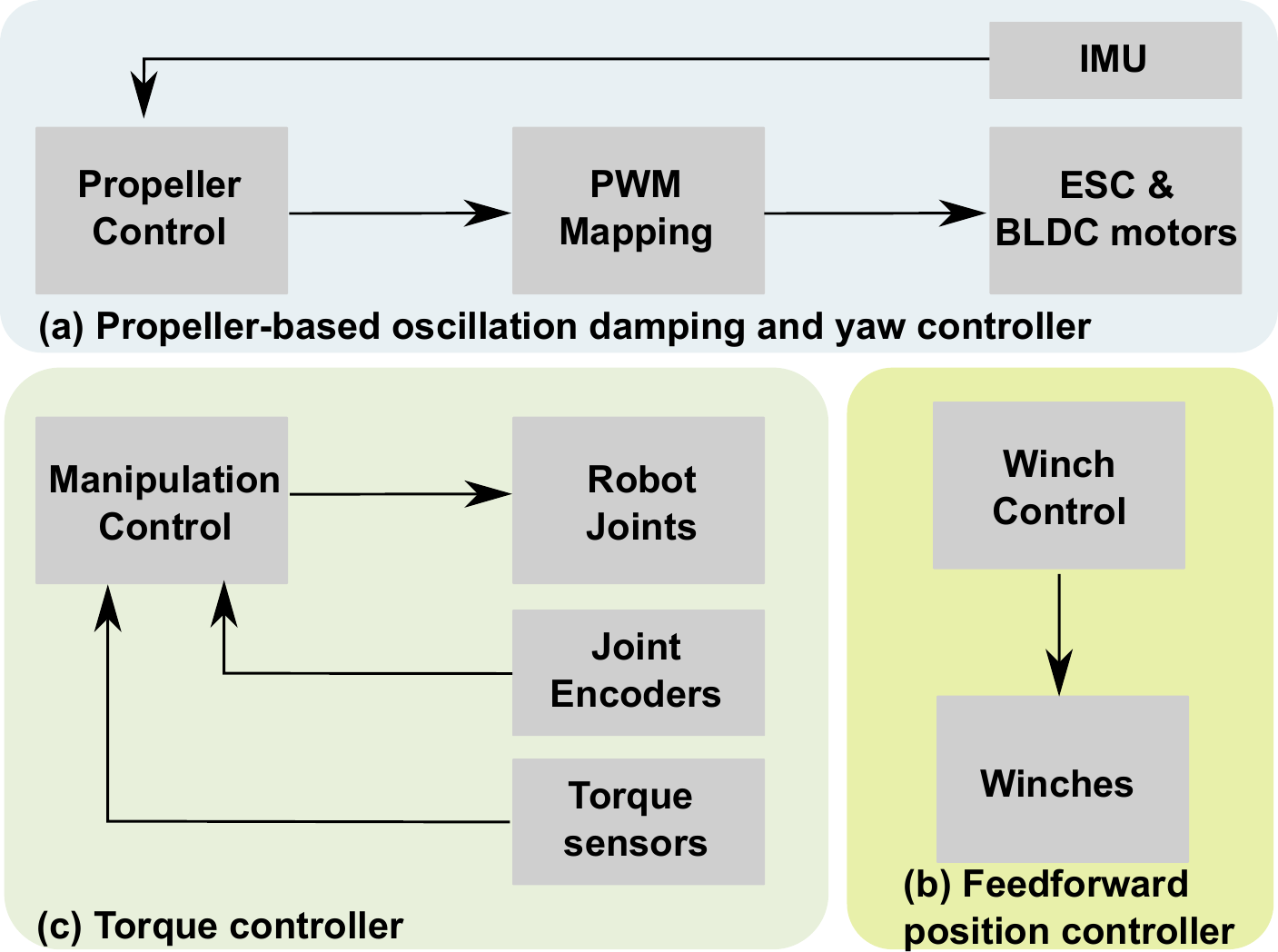}
	}
\end{figure}

\red{The control framework of SAM is depicted in Figure~\ref{fig:appendix:1}. It includes three separate controllers for three sets of actuators. Each of these blocks are to fulfill three different control tasks. The first controller is a propeller based oscillation damping and yaw motion control using IMU as a single main sensor. The main task herein is to damp out oscillations and control yaw motion. Oscillations occur due to the forces and moments caused by the robotic arm, which interacts with the environments. In outdoor settings, severe wind, motion of the carrier, and other external disturbances cause such oscillations. Damping out these undesired motions are to perform precise manipulation tasks with the robotic arm. Similarly, yaw motion controller is to change the orientation of the platform, which can position the manipulator in a convenient pose. To do so, the robot actuation is performed by eight propellers attached to each BLDC motors. The Electronic Speed Controllers (ESCs) regulates BLDC motors to rotate at specific speed. The used sensors are again a single IMU attached to a fixed point of the platform. The control signals are generated by the propeller controller, which is essentially a PID control algorithm. These control signals are mapped to PWM signals per each motors using the known configurations of motors and propellers. The frequency rate is 200Hz in a real-time computer. Secondly, a feedforward position controller is used to control three winches. These winches are connected to cables that suspend the platforms. Maxon motors are used without any feedback signals. The main feature is to control the length of the cables. With these, the pitch and the roll orientation of the platform can be adjusted with slower dynamics. Another advantage is to move the platform up and down, without moving the carrier. Here, a simple feedforward position controller is integrated where desired relative cable length are converted into the motor movements. The frequency rate of the controller is again 200Hz in a real-time computer. Lastly, the robotic arm attached to the platform, is controlled using impedance based torque control algorithm. Torque control is the current golden standards for such robotic arms. The main task here is to perform grasping and manipulation tasks, using the torque control capabilities of the robot. This means that, in teleoperation mode, the robotic arm must follow the command from the human operator, while autonomously taking care of local redundancy of the joints. Joint encoders and torque sensors provide such feedback signals. The manipulator's internal joint torque controller uses sampling rate of 3kHz.}

\subsection{Teleoperation System}

\begin{figure}
	\floatbox[{\capbeside\thisfloatsetup{capbesideposition={right,top},capbesidewidth=4cm}}]{figure}[\FBwidth]
	{\caption{\red{Teleoperation architecture based on Time-Domain Passivity Approach (TDPA). The TDPA approach ensures stability under imperfect communication between the haptic device and the  robot, and consists of two components, namely passivity observer and passivity controller. The proposed telepresence system also includes haptic guidance, robotic perception and VR for 3D visual feedback.}}\label{fig:appendix:2}}
	{\centering
		\includegraphics[width=0.75\textwidth]{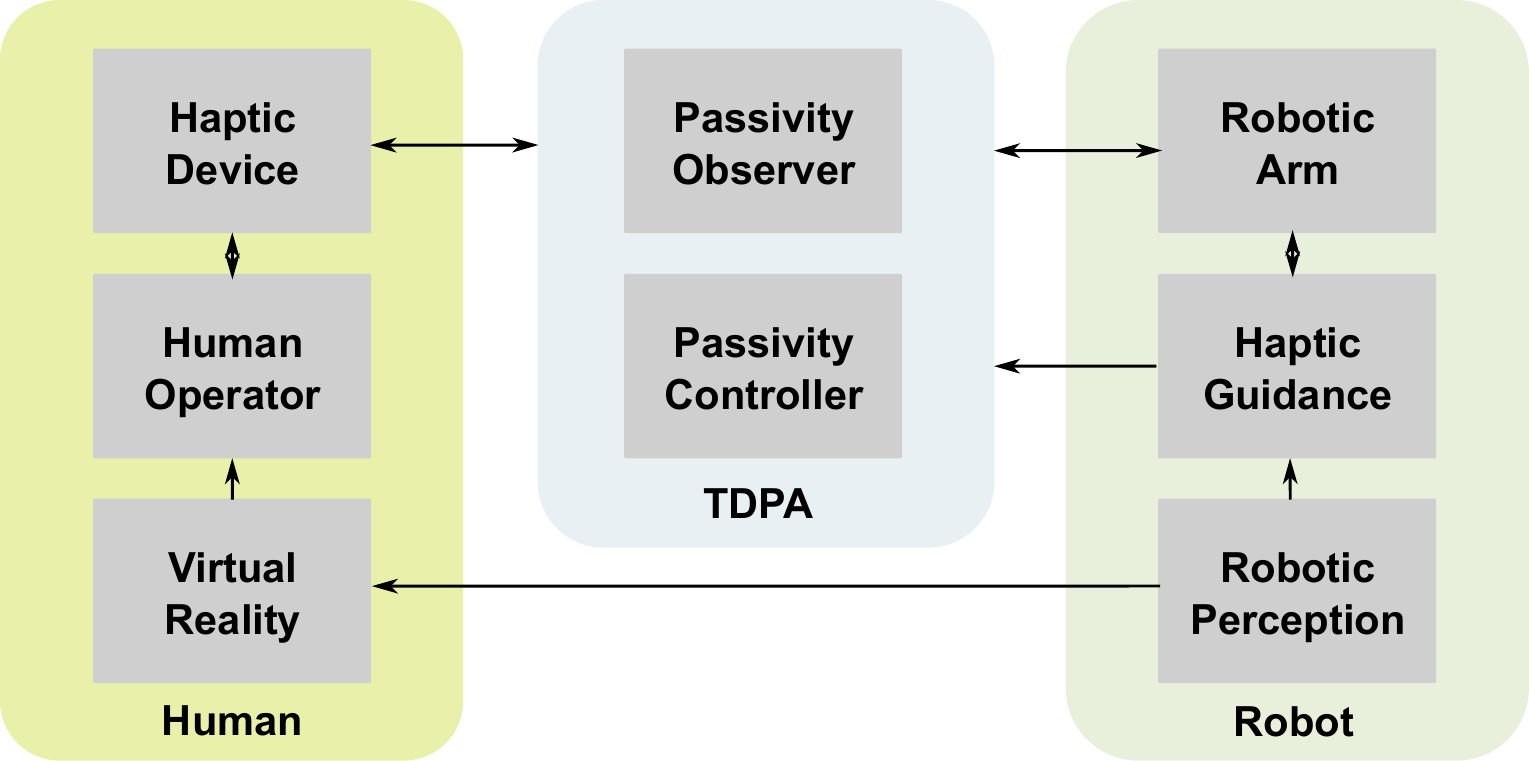}
	}
\end{figure}

\begin{figure}
	\begin{center}
		\includegraphics[width=1\textwidth]{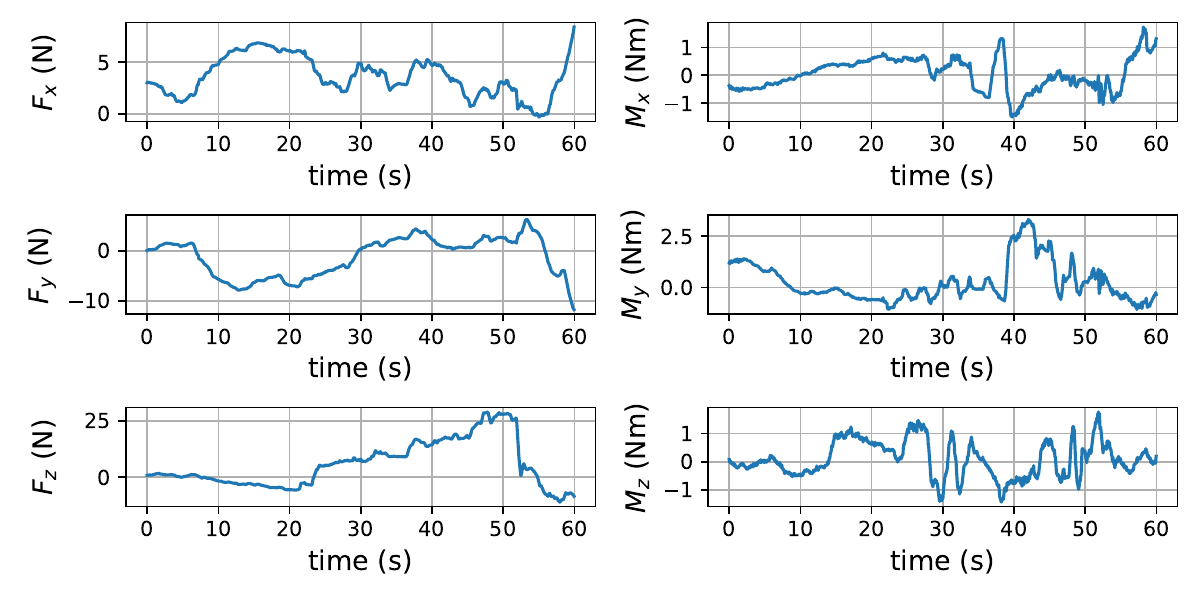}
		\caption{\red{Wrench forces and moments from the peg-in-hole experiments. The poses of the end-effector are alternatively depicted in Figure \ref{fig:night:results}. In $F_z$, between 50s and 60s, a drop in force is observed due to the successful peg-in-hole insertion. }}
		\label{fig:wrench:results1}
	\end{center}
\end{figure}

\begin{figure}
	\begin{center}
		\includegraphics[width=1\textwidth]{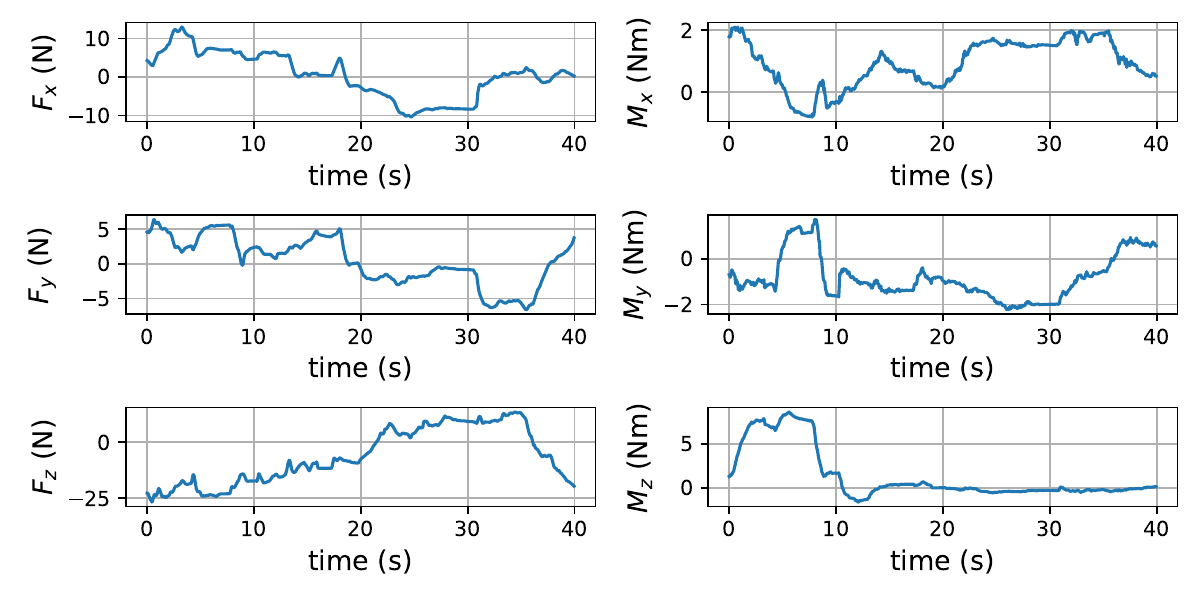}
		\caption{\red{Wrench forces and moments from the pick-and-place experiments. The corresponding poses of the end-effector are depicted in Figure \ref{fig:night:results}. In $Mz$, the haptic guidance activation can be seen, which leads to correction of yaw angle.}}
		\label{fig:wrench:results2}
	\end{center}
\end{figure}

\begin{figure}
	\centering
	\includegraphics[width=1.0\textwidth]{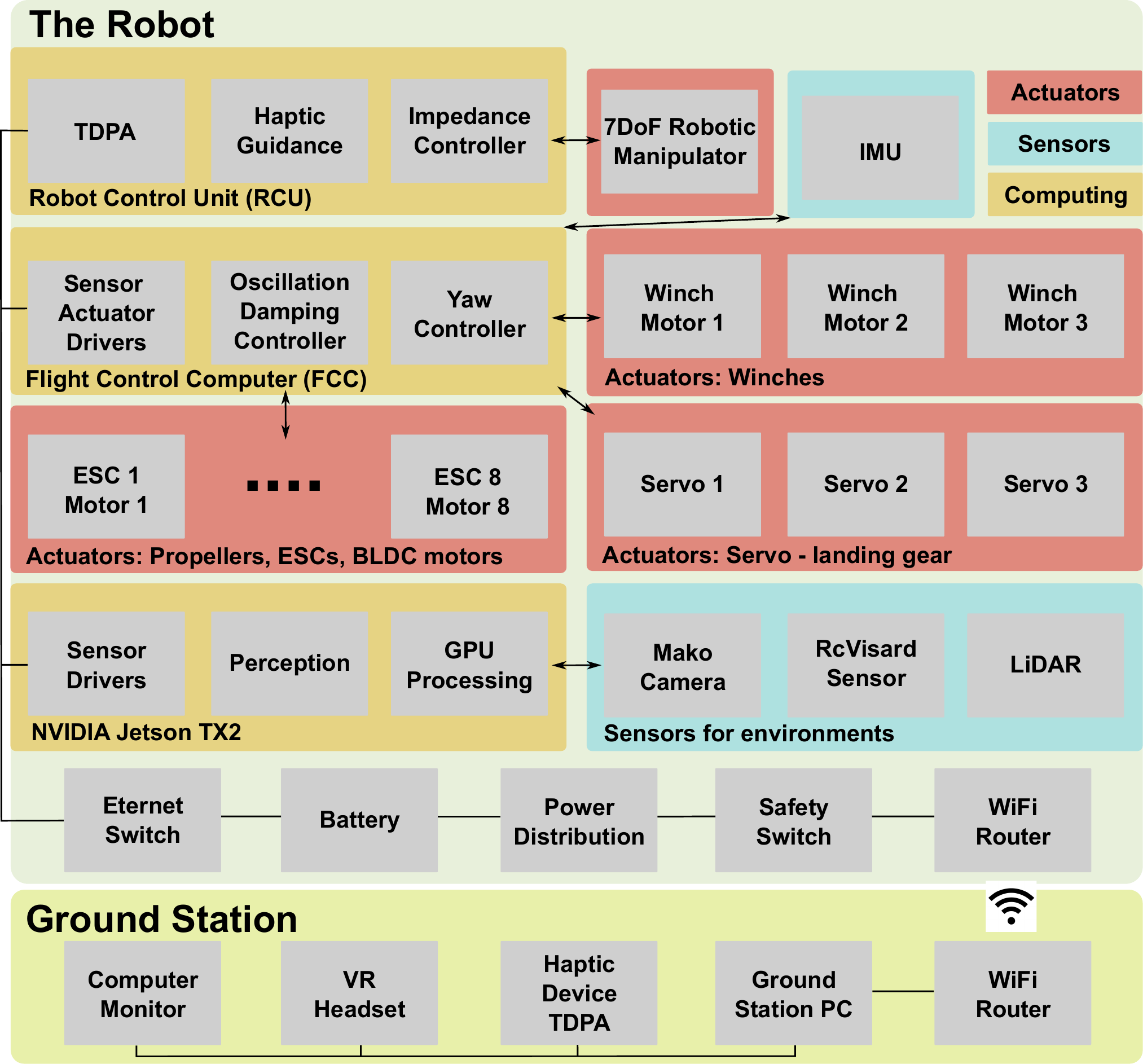}
	\caption{\red{SAM's IT architecture. The broad division is between the ground station and the robot. The ground station hosts haptic device, VR and the human operator. WiFi routers are used to communicate to the robot, which simulates the real teleoperation scenarios with imperfect communication and delays. The robot hosts computers, actuators, sensors and other auxiliary components such as power distribution, battery, etc. SAM has three computers. The robot control unit (RCU) deals with the control of robotic manipulator and hosts real-time linux as its operating system. The flight control computers (FCC) deals with platform control, including propellers, winches and additional servo motors as landing gear. FCC hosts QNX real-time linux system. Lastly, NVIDIA Jetson TX2 is used as a vision processing unit. The TX2 hosts non real-time linux (ubuntu tegra) but has a GPU for deploying the deep learning models.}}
	\label{fig:appendix:3}
\end{figure}

\red{
The used bilateral teleoperation system is depicted in Figure~\ref{fig:appendix:2}. The main challenge is ensuring a stable bilateral teleoperation with force feedback. Instabilities can be caused by time delays, packet loss and jitters, which are characteristics of imperfect communication. 
}

\red{
In the paper, a two channel architecture with time-domain passivity approach is used, which works as follows. The human operator sends both position $v_m(t)$ (velocity analogously) from the haptic device device to the robot at time t. Due to the time-delay T, the robot receives them as $v_{sd}(t) = v_m(t-T)$ where $G_h$ is a scaling factor that can be tuned to match both system dynamics. A local impedance controller then generate reference force $f_{s}(t)$ based on command position. If K terms represent controller gains and $x_s$, $v_{s} \in \mathbb{R}^2$ are the respective feedback signals from the robot which enables position tracking, then the total commanded force can be written as:
}

\begin{equation}
\label{tdpa:eq:1}
    f_s(t) = K_{ds}(v_{sd}(t)-v_s(t)) + K_{ps}(x_{sd}(t)-x_s(t)).
\end{equation}

\red{
The computed force $f_{m0}(t) = G_sf_{s}(t-T)$ and measured forces at the end effector $f_{me}(t) = G_ef_e(t-T)$ are sent back to the haptic device resulting in the force feedback term $f_m(t)$ in Equation~\ref{tdpa:eq:2} where we additionally add a feedforward term with $v_m(t)$. The feedforward terms add transparency to the system and is known to be advantageous over a 2-channel architecture:
}

\begin{equation}
\label{tdpa:eq:2}
    f_m(t) = f_{m0}(t) + f_{me}(t) + K_{dm}v_m(t).
\end{equation}

\red{
As the signals pass through communication channels, time-delays, jitter and packet losses are typically present and can cause instability of overall system. To cope with this issue, we use TDPA which constitutes two components namely Passivity Observer (PO) and Passivity Controller (PC). Briefly speaking, PO monitors the energy flow of a network whereas PC dissipates the energy introduced by the network. A key underlying idea is PC's control law ensures passivity of the system by damping out the energy that is more than the stored amount. Since passivity is a sufficient condition for stability, TDPA ensures stability in trade-off to performance. Therefore, POPCs are placed for delayed signals at the robot side $v_m(t-T)$, and haptic device side $f_{s}(t-T)$. For brevity, let us denote the haptic device signals received and sent as $u_m(k-D)$ and $y_m(k)$, and the robot input and output signals as $u_s(k-D)$ and $y_s(k)$. Here, k is a discrete time and D is a discrete time delay. Then,
}

\begin{equation}
\label{tdpa:eq:3}
u_{m,c}(k) =\left\{\begin{matrix}
u_m(k-D) & \text{if} \ \  W_m(k)> 0 \\ 
u_m(k-D) - \frac{W_m(k)}{T_s y_m^2(k)}y_m(k) & \text{else}
\end{matrix}\right.
\end{equation}

\red{
is the governed control law at haptic device. The same rationale applies at the robot side. In Equation~\ref{tdpa:eq:3}, $T_s$ is the sampling time and $W_m(k)$ is the energy flow at haptic device, which is observed by the PO. In this way, PC modifies the delayed signal so that passivity condition $W_m(k) \geq 0$ for all k is met:
}

\begin{equation}
\begin{array}{lll}
\label{tdpa:eq:4}
\centering
W_m(k)  &=&  E_{\text{s,in}}(k-D)-E_{\text{m,out}}(k)+E_{\text{m,PC}}(k) \\
E_{\text{s,in}}(k-D) &=& E_{\text{s,in}}(k-D-1) + T_s P_{\text{s,in}}(k-D) \\ 
5E_{\text{m,out}}(k) &=& E_{\text{m,out}}(k-1)+T_sP_{\text{m,out}}(k).
\end{array}
\end{equation}

\red{
PO essentially estimates $W_m(k)$ for PCs control law. This is achieved by Equation \ref{tdpa:eq:4} which uses the delayed energy $E_{\text{s,in}}(k-D)$ input from the robot side, the energy exiting at the haptic device side $E_{\text{m,out}}(k)$ and the dissipated energy by PC $E_{\text{m,PC}}(k)$. As the signals being exchanged are velocities $v$ and forces $f$, the energy can be computed by inner products and sampling time. The power contributions should take into account the direction of energy flow. For example, $P_{\text{s,in}} = 0$ if $P_{\text{s,in}} \leq 0)$ and otherwise, $P_{\text{s,in}} = f_s(k)(-v_m(n-D))$. Taking into account time-delays, jitter and packet-loss, TDPA works on energy level and it ensures stability in teleoperation.}

\red{
Besides, the presented perception algorithms are executed in the robot, providing the information about the object poses. This information is feed into the ground station for creating VR. Another usage of perception is haptic guidance via virtual fixtures. Virtual fixtures \citep{Bettini2004} are artificial walls that, by means of force feedback, helps the human operator for high performance task execution. Once the human operator is trying to move outside the artifical walls, certain computed forces are activated and sent to the haptic device through TDPA. This then limits the motion of the human operator by inserting certain forces and moments in the haptic device. Because these artificial walls are obtained from the perception system of the robot, the proposed telepresence system supports the haptic guidance. More details about virtual fixtures and other means of haptic guidances are presented by \citet{hulin2012, Sagardia2018, tubiblio109900}.
}

\subsection{IT Architectures}

\red{
In Figure~\ref{fig:appendix:3}, an overview of the used IT architecture is shown. Broadly, the set-up can be divided into the ground station components and the robot itself. To emulate real industrial scenarios of telepresence robots, the connection between the robot and the ground station is established through a WiFi router. The ground station constitutes of a laptop (Dell Latitude 5591), a haptic device (Force dimension Lambda or DLR SpaceJoystick Ryo), VR headset (Meta occulus) and a computer monitor. VR headset is optionally used. From the robot side, the Flight Control Computer (FCC) is employed, which is a product from the DLR aerial robotic spin-off Elektra UAS. FCC is a QNX based real-time system and contains  a field-programmable gate array (FPGA) based safety switch. FCC is connected to winches, servo motors (Futaba S3152) and ESCs with custom written drivers. Oscillation damping controller, yaw controller and on-and-off of servos are executed within FCC. This modules read data from IMU (Xsens MTi 100-Series). In addition, FCC is also connected to the manual command transmitter via a radio link. The robot control unit (RCU) is based on Kontron KTH81 Flex board and uses real-time linux patch of open suse operating system.  Ethernet for Control Automation Technology (EtherCAT) protocol is used to communication with the robotic arm. We note that EtherCAT is a standardized real-time bus that enables synchronous actuation of all the joint motors. RCU executes TDPA, haptic guidance and impedance controller, while reading joint and torque information from the robotic arm. The last computing module is NVIDIA Jetson TX2 which contains all the sensor drivers, perception software stacks, and other GPU processing modules for running deep learning models. The sensors are all connected via Ethernet interface. For this, Cogswell carrier board is employed, which supports five Ethernet ports with Power of Ethernet (PoE) functionality. With this, Mako camera is easily powered. The carrier board also handles high data throughput from all these sensors. The robot is additionally equipped with safety switch for the robotic arm, power distribution system and battery. All the computers and WiFi routers are connected via an Ethernet switch (Netgear GS105). The communication between the ground station and the robot is through the point-to-point communication channel by opening an access point. Ubiquiti Bullet M5 is employed for the access point.}


\subsubsection*{Acknowledgements} 

The authors would like to thank Michael Panzirsch and Nicolai Bechtel for the support regarding Lambda, Nidhish Raj for participating in the user validation study, and other members of the flying robots, like Min Jun Kim and Yura Sarkisov, and Elektra UAS team for the support. Special thanks to Martin Schuster and Lukas Meyer for providing detailed feedback on the manuscript. This work is also supported by the Helmholtz Association’s Initiative and Networking Fund (INF) under the Helmholtz AI platform grant agreement (ID ZT-I-PF-5-1) and by the European Commission under the contract 644271 EU2020 AEROARMS and 824990 EU2020 RIMA.

\bibliographystyle{apalike} 
\bibliography{bibliography}

\end{document}